%% file: article.tex
\documentclass[review,fleqn]{cas-sc}

\usepackage[numbers]{natbib}
\usepackage{subfigure}
\usepackage{setspace}
\usepackage{soul}
\usepackage{color}
\usepackage{booktabs}
\usepackage{tabularx}
\usepackage{amsmath}
\usepackage{subcaption}
\usepackage{algorithm}
\usepackage[acronym]{glossaries}
\usepackage{algpseudocode}
\usepackage{todonotes}

\input{acronym}

\def\tsc#1{\csdef{#1}{\textsc{\lowercase{#1}}\xspace}}
\tsc{WGM}
\tsc{QE}
\tsc{EP}
\tsc{PMS}
\tsc{BEC}
\tsc{DE}

\begin{document}
\let\WriteBookmarks\relax
\def\floatpagepagefraction{1}
\def\textpagefraction{.001}
\shorttitle{Enhancing Maritime Situational Awareness through End-to-End Onboard Raw Data Analysis}
\shortauthors{R. Del Prete et~al.}

\title [mode = title]{Enhancing Maritime Situational Awareness by End-to-End Onboard Raw Data Analysis}                      
\tnotemark[1,2]



\author[1,2]{Roberto {Del Prete}*}[type=editor,
                        auid=000,bioid=1,
                        prefix=,
                        role=,
                        orcid=0000-0003-0810-4050]
\author[1,3]{Manuel {Salvoldi}}[type=editor,
                        auid=000,bioid=1,
                        prefix=,
                        role=,
                        orcid=0000-0001-5810-6156]
\author[1,4]{Domenico {Barretta}}[type=editor,
                        auid=000,bioid=1,
                        prefix=,
                        role=,
                        orcid=0000-0001-0000-0000]
\author[1]{Nicolas {Longépé}}[type=editor,
                        auid=000,bioid=1,
                        prefix=,
                        role=,
                        orcid=0000-0001-0000-0000]
\author[1, 5]{Gabriele {Meoni}}[type=editor,
                        auid=000,bioid=1,
                        prefix=,
                        role=,
                        orcid=0000-0001-9311-6392]
\author[3]{Arnon {Karnieli}}[type=editor,
                        auid=000,bioid=1,
                        prefix=,
                        role=,
                        orcid=0000-0001-8065-9793]

\author[2]{Maria Daniela {Graziano}}[type=editor,
                        auid=000,bioid=1,
                        prefix=,
                        role=,
                        orcid=0000-0001-0000-0000]
\author[2]{Alfredo {Renga}}[type=editor,
                        auid=000,bioid=1,
                        prefix=,
                        role=,
                        orcid=0000-0002-1236-0594]


\credit{Conceptualization of this study, Methodology, Software}

\affiliation[1]{organization={$\Phi$-Lab, European Space Agency},
                addressline={Via Galileo Galilei 1}, 
                city={Frascati},
                postcode={00044}, 
                state={Rome},
                country={Italy}}


\affiliation[2]{organization={Department of Industrial Engineering, University of Naples Federico II},
                addressline={P.le Vincenzo Tecchio 80}, 
                postcode={80125}, 
                city={Napoli},
                country={Italy}}

\affiliation[3]{organization={Ben-Gurion University of the Negev},
                addressline={Sede Boker Campus}, 
                postcode={8499000}, 
                city={Negev},
                country={Israel}}

\affiliation[4]{organization={Department of Engineering, University of Campania ``L. Vanvitelli''},
                addressline={Via Roma 29}, 
                postcode={81031}, 
                city={Caserta},
                country={Italy}}

\affiliation[5]{organization={Advanced Concepts and Studies Office, European Space Agency},
            addressline={Via Galileo Galilei 1}, 
                city={Frascati},
                postcode={00044}, 
                state={Rome},
                country={Italy}}


\cortext[cor1]{Corresponding author}


\begin{abstract}
Satellite-based onboard data processing is crucial for time-sensitive applications requiring timely and efficient rapid response. Advances in edge artificial intelligence are shifting computational power from ground-based centers to on-orbit platforms, transforming the "sensing-communication-decision-feedback" cycle and reducing latency from acquisition to delivery.
The current research presents a framework addressing the strict bandwidth, energy, and latency constraints of small satellites, focusing on maritime monitoring. The study contributes three main innovations.
Firstly, it investigates the application of deep learning techniques for direct ship detection and classification from raw satellite imagery. By simplifying the onboard processing chain, our approach facilitates direct analyses without requiring computationally intensive steps such as calibration and ortho-rectification.
Secondly, to address the scarcity of raw satellite data, we introduce two novel datasets, VDS2Raw and VDV2Raw, which are derived from raw data from Sentinel-2 and Vegetation and Environment Monitoring New Micro Satellite  (VENµS) missions, respectively, and enriched with Automatic Identification System (AIS) records. 
Thirdly, we characterize the tasks' optimal single and multiple spectral band combinations through statistical and feature-based analyses validated on both datasets.
In sum, we demonstrate the feasibility of the proposed method through a proof-of-concept on CubeSat-like hardware, confirming the models' potential for operational satellite-based maritime monitoring. 
\end{abstract}


\begin{keywords}
Sentinel-2 \sep VENµS\sep Raw MultiSpectral Data\sep Vessel Detection\sep Vessel Classification\sep Onboard Processing
\end{keywords}

\maketitle

\input{sections/01_intro}
\input{sections/02_data}

\input{sections/03_method}
\input{sections/04_results}
\input{sections/05_towards}

\input{sections/06_conclusion}

\bibliographystyle{unsrt}
\bibliography{article.bib}





\end{document}

%% file: acronym.tex
\newacronym{AI}{AI}{Artificial Intelligence}
\newacronym{AIS}{AIS}{Automatic Identification System}
\newacronym{CNN}{CNN}{Convolutional Neural Network}
\newacronym{DN}{DN}{Digital Number}
\newacronym{CPU}{CPU}{Central Processing Unit}
\newacronym{CSC}{CSC}{Coarse Spatial Coregistration}
\newacronym{CV}{CV}{Cross-Validation}
\newacronym{EO}{EO}{Earth Observation}
\newacronym{ED}{ED}{Euclidean Distance}
\newacronym{FLOAT16}{FLOAT16}{16-bit Floating Point}
\newacronym{L0}{L0}{Level-0}
\newacronym{L1A}{L1A}{Level-1A}
\newacronym{L1B}{L1B}{Level-1B}
\newacronym{L1C}{L1C}{Level-1C}
\newacronym{LEO}{LEO}{Low Earth Orbit}
\newacronym{MCC}{MCC}{Matthew Correlation Coefficient}
\newacronym{ML}{ML}{Machine Learning}
\newacronym{NCS}{NCS}{Neural Comput Stick}
\newacronym{NCS2}{NCS2}{Neural Comput Stick 2}
\newacronym{SHAVE}{SHAVE}{Streaming Hybrid Architecture Vector Engine}
\newacronym{THRawS}{THRawS}{Thermal Hotspots in Raw Sentinel-2 images}
\newacronym{VPU}{VPU}{Vision Processing Unit}
\newacronym{FPGAs}{FPGAs}{Field Programmable Gate Arrays}
\newacronym{GPUs}{GPUs}{Graphics Processing Units}
\newacronym{MDA}{MDA}{Maritime Domain Awareness}
\newacronym{MSA}{MSA}{Maritime Situational Awareness}
\newacronym{DL}{DL}{Deep Learning}
\newacronym{ESA}{ESA}{European Space Agency}
\newacronym{FPS}{FPS}{Frames Per Second}
\newacronym{S-2}{S-2}{Sentinel-2}
\newacronym{MSI}{MSI}{MultiSpectral Instrument}
\newacronym{ROI}{ROI}{Region of Interest}
\newacronym{VSSC}{VSSC}{VENµS SuperSpectral Camera}
\newacronym{MTF}{MTF}{Modulation Transfer Function}
\newacronym{SNR}{SNR}{Signal-to-Noise Ratio}
\newacronym{PSF}{PSF}{Point Spread Function}
\newacronym{ISA}{ISA}{Israeli Space Agency}
\newacronym{IoU}{IoU}{Intersection over Union}
\newacronym{HOG}{HOG}{Histogram of Oriented Gradients}
\newacronym{PCC}{PCC}{Pearson Correlation Coefficient}
\newacronym{MMSI}{MMSI}{Maritime Mobile Service Identity}
\newacronym{RADC}{RADC}{Residual-Attention-Dilated-Convolution}
\newacronym{CAM}{CAM}{Channel Attention Module}
\newacronym{SAHI}{SAHI}{Slicing Aided Hyper Inference}
\newacronym{NMS}{NMS}{Non-Maximum Suppression}
\newacronym{TOA}{TOA}{Top Of Atmosphere}

\newacronym{UTC}{UTC}{Coordinated Universal Time}
\newacronym{CCD}{CCD}{Charge-Coupled Device}
\newacronym{TDI}{TDI}{Time Delay Integration}

%% file: sections/01_intro.tex
\section{Introduction}\label{sec:intro}
Considering the enormous impact on maritime applications, the efficient and timely recognition of vessels using \gls{EO} satellite imagery is imperative for scenarios requiring rapid response. Applications such as traffic and environmental monitoring, emergency search and rescue operations, and detecting illegal fishing or smuggling require rapid, low-latency responses to be effective.
The traditional data processing chains based on the classical bent-pipe approaches \cite{furano2020towards,GiuffridaPhiSat} face significant challenges in addressing these needs due to several inherent limitations, such as the increased latency for data download, 
ground station availability and high-level product calculation, thereby incurring substantial delays from image acquisition to information delivery \cite{meoni2024unlocking}. 
Additionally, the need to transmit satellite data to Earth strains the available communication bandwidth \cite{furano2020towards,GiuffridaPhiSat}.
\newline
AI can enable real-time data processing, reducing the need for large data transfers to Earth and accelerating response times to various events \cite{wijata2023taking}.
A growing body of research is focused on leveraging \gls{AI} onboard satellites to retrieve actionable information for latency-sensitive applications quickly. This includes natural disaster response \cite{mateo-garcia_towards_2021,diana2021oil,meoni2024unlocking} and the detection of anomalies or targets in localized areas \cite{ruzicka2022ravaen,del2023first}. The increasing interest in this area is further demonstrated by the rising number of companies and research institutes worldwide developing advanced edge \gls{AI} avionics subsystems for CubeSats \cite{10597570}.
Recent missions, such as $\Phi$Sat-2 \cite{melega2023implementation}, Intuition-1 by KP Labs, CogniSAT-6U by Ubotica \cite{rijlaarsdam2023autonomous}, and Kanyini by SmartSat CRC \cite{kanyini}, showcase the growing commitment to AI-enabled satellite technology. These missions utilize \gls{AI}-enabled processing units, demonstrating the significant advantages of \gls{AI} in enhancing real-time maritime surveillance and other critical applications. This trend underscores the pivotal role of onboard \gls{AI} as the key to revolutionizing maritime monitoring and response, providing faster, more accurate, and more efficient operations in the ever-challenging maritime domain.
\newline
Past satellite missions, such as $\Phi$-Sat-1 \cite{GiuffridaPhiSat} and HYPSO-1 \cite{danielsen2021self}, relied on extensive pre-processing workflows, including geometric and radiometric corrections, to prepare data for onboard \gls{ML} applications. $\Phi$-Sat-1, for example, implemented a \gls{CNN} on an Intel® Movidius™ Myriad™ 2 \gls{VPU}, processing selected hyperspectral bands after creating the hyperspectral data cube and performing band-to-band spatial registration. Similarly, HYPSO-1 employed a hyperspectral payload to monitor ocean color, requiring onboard image processing that involved linear radiometric and geometric corrections. 
\newline
Although current research is advancing towards novel technologies, particularly in \gls{DL}, few attention has been dedicated to the pre-processing stage, which continues to mimic traditional ground-based workflows.
%
The reliance on such demanding schemes necessitates specialized hardware on orbit which increases payload complexity and resource utilzation. Therefore, to optimize onboard resource usage while providing real-time actionable information, it is essential to bypass all unnecessary pre-processing steps, realising models that can analyse raw imagery. 
Tackling this issue poses several challenges, including hardware and energy limitations, resource efficiency of \gls{ML} models, but also the scarcity of available raw data datasets.
\newline
The very first direct attempt to process raw data end-to-end for onboard machine learning was made during the ``OPS-SAT Case'' competition \cite{derksen2021few,meoni2024ops}, hosted on the \gls{ESA} Kelvins platform, which focused on few-shot learning for satellite applications. This effort represented a significant shift from previous missions, directly tackling the challenges of onboard raw data processing. Building on this, Meoni et al. \cite{meoni2023thraws} developed a methodology for creating raw datasets using \gls{S-2} imagery, providing the first raw dataset of thermal hotspots. 
Continuing this trend, Del Prete et al. \cite{del2023first} introduced VDS2Raw, a dataset specifically designed for vessel detection using raw \gls{S-2} data, offering a comparative analysis of various \gls{DL} techniques for the task.
$\Phi$Sat-2 \cite{melega2023implementation} offers the possibility of processing \gls{L1A} data on board thanks to an advanced data processing framework for generating multispectral and panchromatic imagery at three distinct levels
\footnote{The \gls{L1A} product provides \gls{TOA} radiance without geo-referencing or band alignment, while Level 1B enhances this with precise geo-referencing and band alignment. Level 1C delivers \gls{TOA} reflectance with fine geo-referencing and band alignment, though it lacks orthorectification.}.
Emphasizing on this innovative apporach of raw data exploitation, the \textit{Orbital AI} challenge \cite{longepe2024simulation} was the first to explore onboard AI applications with raw satellite data simulated.
\newline
This manuscript builds upon the aforementioned works and makes several novel key contributions:
\begin{enumerate}
    \item \textbf{End-to-End Workflow:} This work presents the first fully integrated end-to-end workflow for onboard vessel identification (detection and classification), streamlining the process from data acquisition to information delivery, thereby reducing latency and improving the efficiency of maritime surveillance operations.
    
    \item \textbf{Dataset Creation:} Two new datasets of raw, uncalibrated multispectral data for vessel detection \& classification have been created, incorporating additional \gls{AIS} information and a novel annotation format that includes both bounding boxes and \gls{AIS} records. These datasets are built from two different EO space missions with different characteristics 
    
    \item \textbf{Spectral Band Analysis:} Through the application of DL techniques on the developed datasets, the study identifies the most useful raw spectral bands across the spectrum for vessel detection and classification, providing insights into the performance across different geographic areas, sensors, and resolutions.
    
    \item \textbf{Onboard Implementation:} A proof-of-concept for onboard implementation is demonstrated using an \gls{AI} edge device with flight heritage, such as CogniSAT-6 \cite{rijlaarsdam2023autonomous, rijlaarsdam2024next}. This deployment, tested under varying sensor operating conditions, confirms the feasibility of real-time and efficient maritime surveillance.
\end{enumerate}
The remainder of the paper is organized as follows. 
Section 2 details the data creation and curation strategy for the two raw multispectral datasets designed for vessel detection and classification, enriched with AIS information.  
Section 3 discusses the proposed methodology, focusing on the cascaded application of coregistration and DL-based detection techniques.  
Section 4 presents the results obtained from the developed datasets, highlighting the generalization and applicability of the proposed approach to several multispectral bands.  
Section 5 is dedicated to the onboard implementation, showcasing a proof-of-concept demonstration by representative hardware. 
Finally, Section 6 provides a discussion of the results and concludes the paper.

%% file: sections/02_data.tex
\section{Datasets Creation Strategy} \label{sec:data}

This section details the development of raw multispectral datasets. The two selected space missions are initially presented, i.e., Sentinel-2 and VENµS. Then, inherent characteristics and challenges associated with raw data are examined, and a corresponding solution is proposed. Subsequently, the data acquisition and labelling strategy are reported along with the characteristics of the datasets derived from the two imagers. 
Concluding the section, the problem of matching AIS records with remote sensing data is addressed.

    \subsection{The Sentinel-2 and VENµS Missions}
\begin{figure}[h] 
            \centering
            \includegraphics[width=0.85\textwidth]{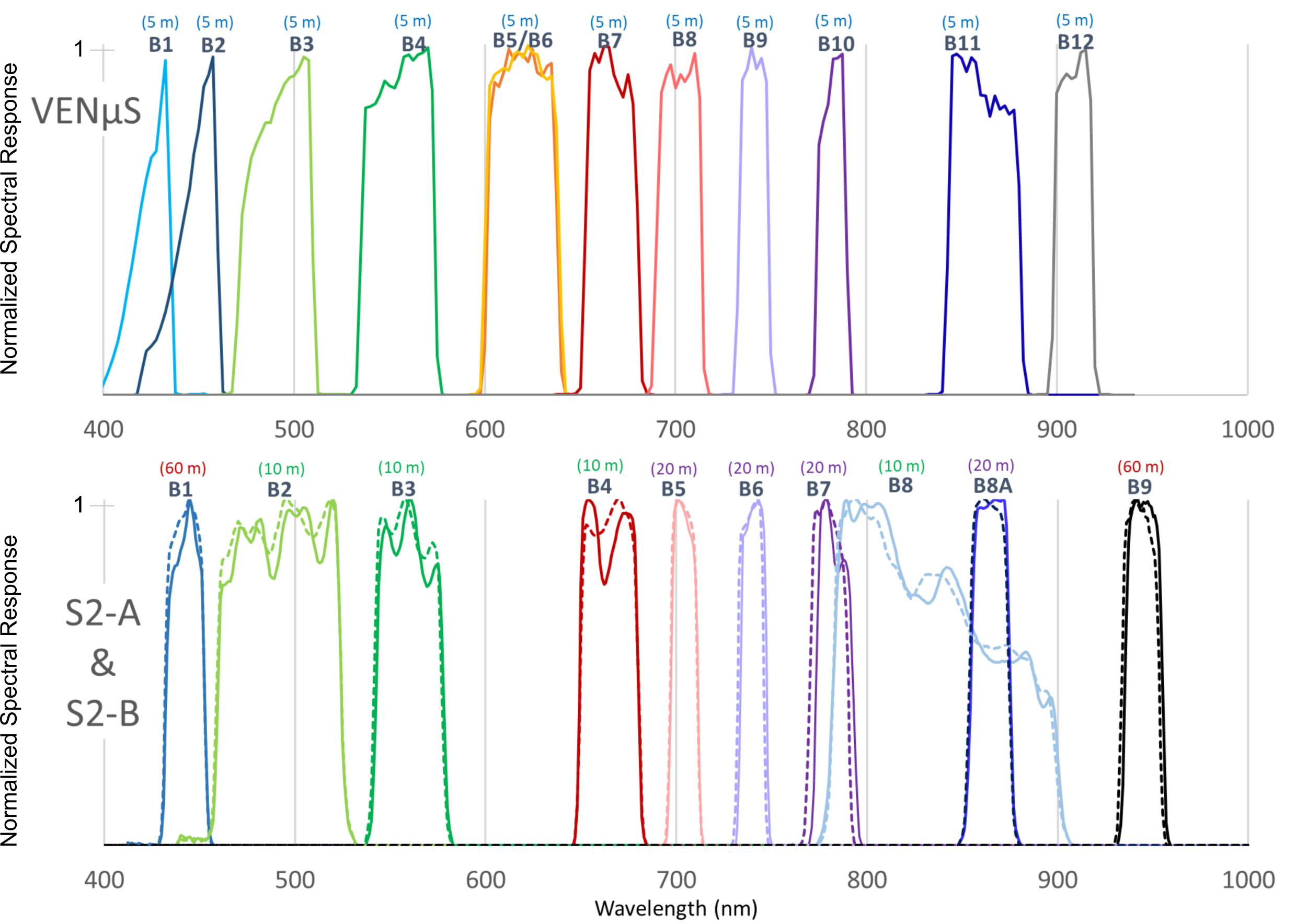}
            \caption{Graphical comparison of the normalized spectral responses between the \gls{MSI} on the VENµS satellite (top) and the \gls{S-2} (bottom) highlighting key differences in their spectral band coverage. Specifically, the VENµS bands $B_i$ ($i=1,5,6,12$) do not have corresponding bands in the \gls{S-2} imager. Conversely, bands $B_8$ and $B_9$ of \gls{S-2} are absent in VENµS, indicating that each \gls{MSI} is optimized for different spectral features.}
            \label{fig:VENuS_S2_bands}
        \end{figure}

        The Sentinel-2 mission, an integral component of the European Union's Copernicus program, comprises three satellites: Sentinel-2A, which was launched in June 2015, Sentinel-2B, launched in March 2017, and Sentinel-2C, recently deployed on 5 September 2024. 
        These satellites are designed for high-resolution, multispectral imaging of Earth’s land and coastal areas. Sentinel-2 satellites operate in a sun-synchronous orbit with a 290 km swath width and a revisit time of 5 days at the Equator. Each satellite is equipped with a \gls{MSI} that captures 13 spectral bands at different spatial resolutions: four bands at 10 meters ($B_2$, $B_3$, $B_4$, $B_8$), six bands at 20 meters ($B_5$, $B_6$, $B_7$, $B_{8A}$, $B_{11}$, $B_{12}$), and three bands at 60 meters ($B_1$, $B_9$, $B_{10}$). These capabilities allow for detailed observation of vegetation, soil, water cover, and changes in land use. Data products from \gls{S-2} are available at different processing levels: Level-1C (top-of-atmosphere reflectance), Level-2A (bottom-of-atmosphere reflectance), and higher-level products. The MSI mounts 12 detectors in a staggered configuration, chosen for achieving an high swathwidth. Each detector produces a granule containing the aforementioned spectral bands. 
        As defined in \cite{meoni2024unlocking}, \gls{L0} data refers to the sensor-acquired information that has been equalized and compressed onboard the satellite before being transmitted to the ground. After downlinking, this data is decompressed and supplemented with metadata, but it remains unprocessed in terms of geometric and radiometric correction\footnote{Further details are available in the \gls{S-2} Product Specification Document (PSD v15): \url{https://sentinel.esa.int/documents/d/sentinel/s2-pdgs-cs-di-psd-v15-0}}. We refer to this data as raw data, thus, representing the earliest stage of the processing chain after decompression, preceding further refinement steps such as co-registration, radiometric calibration, ortho-rectification, required for the generation of higher level products.
        The only publicly available datasets of raw \gls{S-2} data are those provided in our previous works \cite{meoni2024unlocking,del2023first}, as \gls{L0} products are generally not distributed to the public. 
        \newline
        The VENµS mission, a collaboration between the Israel Space Agency (ISA) and the French Space Agency (CNES), has been in sun-synchronous LEO since August 2017, acquiring multispectral imagery every two days over 120+ sites at 5-meter resolution, covering the visible (VIS) to near-infrared (NIR) spectrum for environmental and scientific research.
        The VENµS \gls{MSI}, i.e., the \gls{VSSC}, provide observations with high spatial, temporal, and spectral resolutions \cite{SALVOLDI202233}, \cite{HERRMANN20112141}, \cite{BERGSMA2021112469}. The \gls{VSSC} is a push broom imager comprising a catadioptric objective (0.25 m in diameter), a focal plane (1.75 m in length) assembly with narrow-band filters, and four detector units with three separate charge-coupled device-time delay integration (\gls{CCD}-\gls{TDI}) arrays each one containing the sensing capability for three spectral bands. The first detector (A) includes bands $B_1$, $B_2$, and $B_5$; the second (B) includes bands $B_{10}$, $B_{11}$, and $B_{12}$; the third (C) includes bands $B_7$, $B_8$, and $B_9$; and the fourth (D) includes bands $B_3$, $B_4$, and $B_6$. This configuration results in a pre-scheduled capture of all 12 spectral bands during each multispectral image acquisition. 
        Three product levels are available: \gls{L1C}, Level-2A (L2), and Level-3 (L3, 10-day L2 composites) \cite{Dick2022}. 
        \gls{L0} products are archived but have never been disseminated to end-users. 
        Nevertheless, the Remote Sensing Laboratory\footnote{https://karnieli-rsl.com/} at Ben Gurion University of the Negev processes, archives, and disseminates products for three Israeli sites. These latter have been exploited to develop two novel comprehensive datasets detailed in the upcoming section of this manuscript. 
        \newline
        Finally, the Figure \ref{fig:VENuS_S2_bands} compares the spectral responses of the \gls{MSI}s of VENµS and \gls{S-2}, highlighting their similarities and differences.
    \subsection{Raw Data Characteristics}

\begin{figure}[b]
            \centering
            \includegraphics[width=\linewidth, trim={1.25cm 23.5cm 1.25cm 1.25cm}, clip]{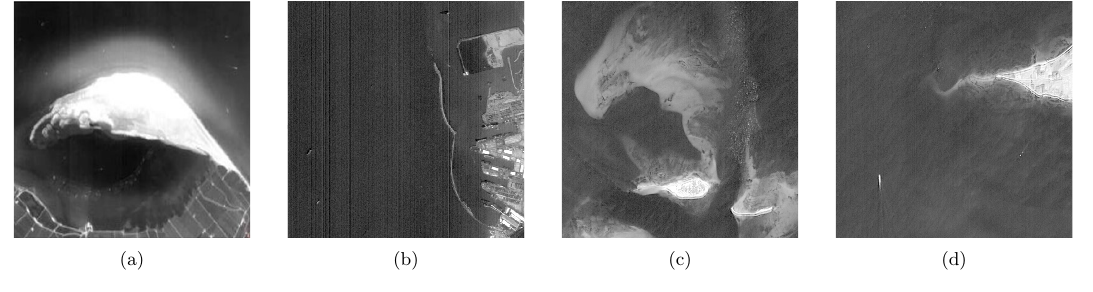}
            \caption{
            (a) Unequalized VENµS \gls{L0} image of the Ebro Delta, showing significant stray light artifacts in band $B_2$, which affects the visibility of underlying features.
            (b) Striping noise observed in band $B_2$ of the VENµS \gls{L0} image captured over the port of Ashdod, illustrating the impact of lack of calibration on coastal imagery.
            (c) Radiometric noise present in band $B_3$ of the \gls{S-2} \gls{L0} image over an algal bloom area, highlighting the challenges of detecting subtle biological features amidst sensor noise.
            (d) Radiometric noise detected in band $B_3$ of the \gls{S-2} \gls{L0} image over the Danish Fjord, demonstrating the effects of sensor-induced noise on the accurate interpretation of aquatic environments.
            }
            \label{fig:L0_examples}
        \end{figure}

        As discussed earlier, most of literature focuses on methods applied to high-level products, due to the limited use and availability of raw data.
        Consequently, approaches for handling raw data remain relatively new and underdeveloped, as working with raw data introduces unique and often uncharted challenges. In this section, we outline these challenges and propose effective solutions to address them.
        \newline
        It is essential to recognize that the \gls{L0} bands represent raw, decompressed data in its most fundamental form. As such, they do not undergo the standard ground segment processing pipeline, which includes computationally-intensive tasks such as calibration, ortho-rectification, and georeferencing; processing required to generate an L1C product.
        Indeed, they exhibit various forms of radiometric noise and artifacts (Figure \ref{fig:L0_examples}). In the case of VENµS, this noise is particularly pronounced, including effects such as stray light \cite{gamet2019measuring} or signal non-uniformity across the camera detectors' pixels and fixed pattern noise (stripe noise). 
        This spatially coherent kind of noise is typically found in multispectral data \cite{liu2023automatic} and arises from improper correction of the \gls{CCD} response function, owing to unexpected temperature changes on the photoelectric system \cite{carfantan2009statistical}. 
        Lastly, an unusual artifact, termed ``radiometric spike'' was identified in VENµS images by \cite{Dick2018}. These spikes affect hundreds of pixels across all spectral bands, both in-flight and during ground measurements, showing an offset variation dependent on incoming radiance. The spikes’ amplitude and location vary by spectral band, but their shape remains consistent. Each spike influences four consecutive pixels oppositely based on their position in the left or right register, and their locations are stationary between ground and in-flight observations. 
        The Figure \ref{fig:L0_examples}(a-b) clearly shows the artifacts and noise affecting VENµS imagery.
        \newline
        For what concerns \gls{S-2} images, the data do not present the same level of noise and artifacts of VENµS \gls{MSI} as the Figure \ref{fig:L0_examples}(c-d) highlights. 
        This is due to onboard equalization, aimed at enhancing the compression rates, that takes into account dark signal variations and inter-pixel offsets \cite{gatti2015sentinel,gascon}. 
        \newline
        While noise and artifacts can be effectively managed by the robust generalization capabilities of \gls{DL} methods, the lack of band alignment poses more significant challenges. Specifically, (a) models may lose crucial spatial information, potentially undermining the accuracy of the analysis, and (b) annotations, especially at the pixel and box levels, demand meticulous attention to ensure consistent accuracy across all spectral bands. The ability of \gls{DL} to handle noise and artifacts is crucial, but precise band alignment remains critical for maintaining spatial coherence and annotation integrity.
        The issue becomes particularly pronounced for small objects like vessels, where even minor discrepancies, such as 1-pixel shift, can substantially impact the model performance (refer to Section \ref{sec: metrics}). 

    \subsection{Data Acquisition and Labelling}\label{sec:Vessel_datasets}

\begin{figure}[t]
            \centering
            \includegraphics[width=0.9\linewidth]{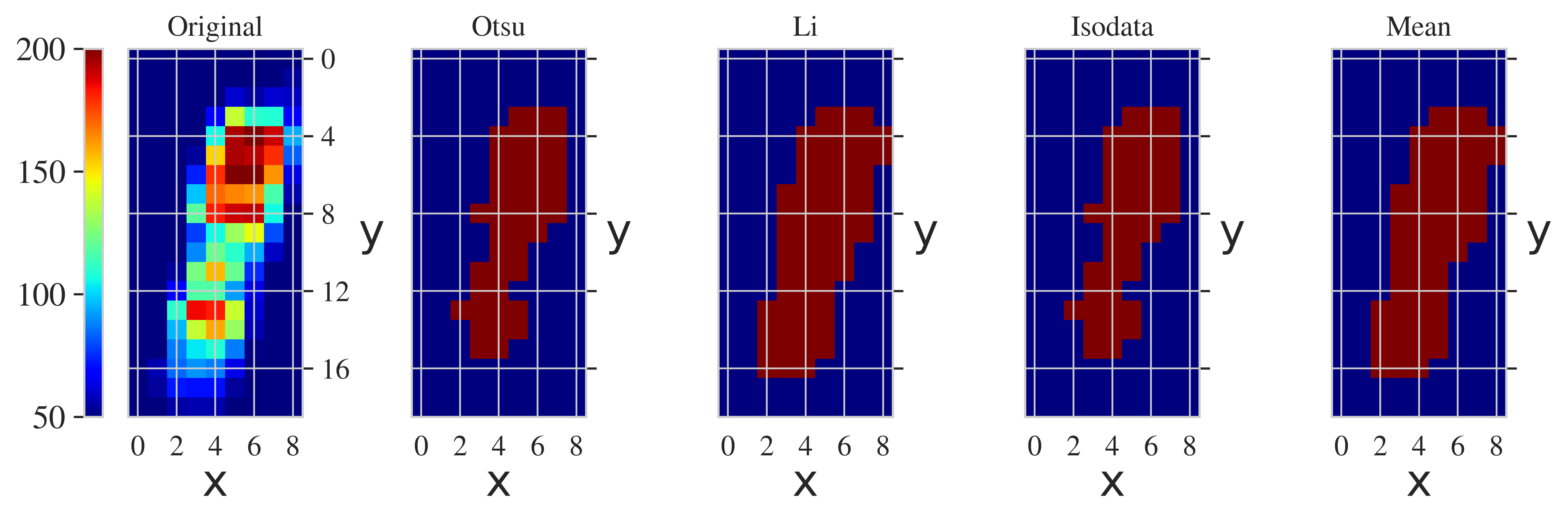}
            \caption{Bounding box fitting of a label in the $B_6$ band (VENµS \gls{MSI}). Image gathered over the port of Ashdod on 2020/01/06. The vessel, identified by \gls{MMSI} 636019532, is a \textit{Container Ship} with a length of 159.8 meters and a width of 24.8 meters. The ship was located at a longitude of 34.58744 and a latitude of 31.8442 at the time of the capture.}
            \label{fig:thresholding}
        \end{figure}

\begin{figure*}[!h]
        \centering
        \includegraphics[width=0.9\linewidth]{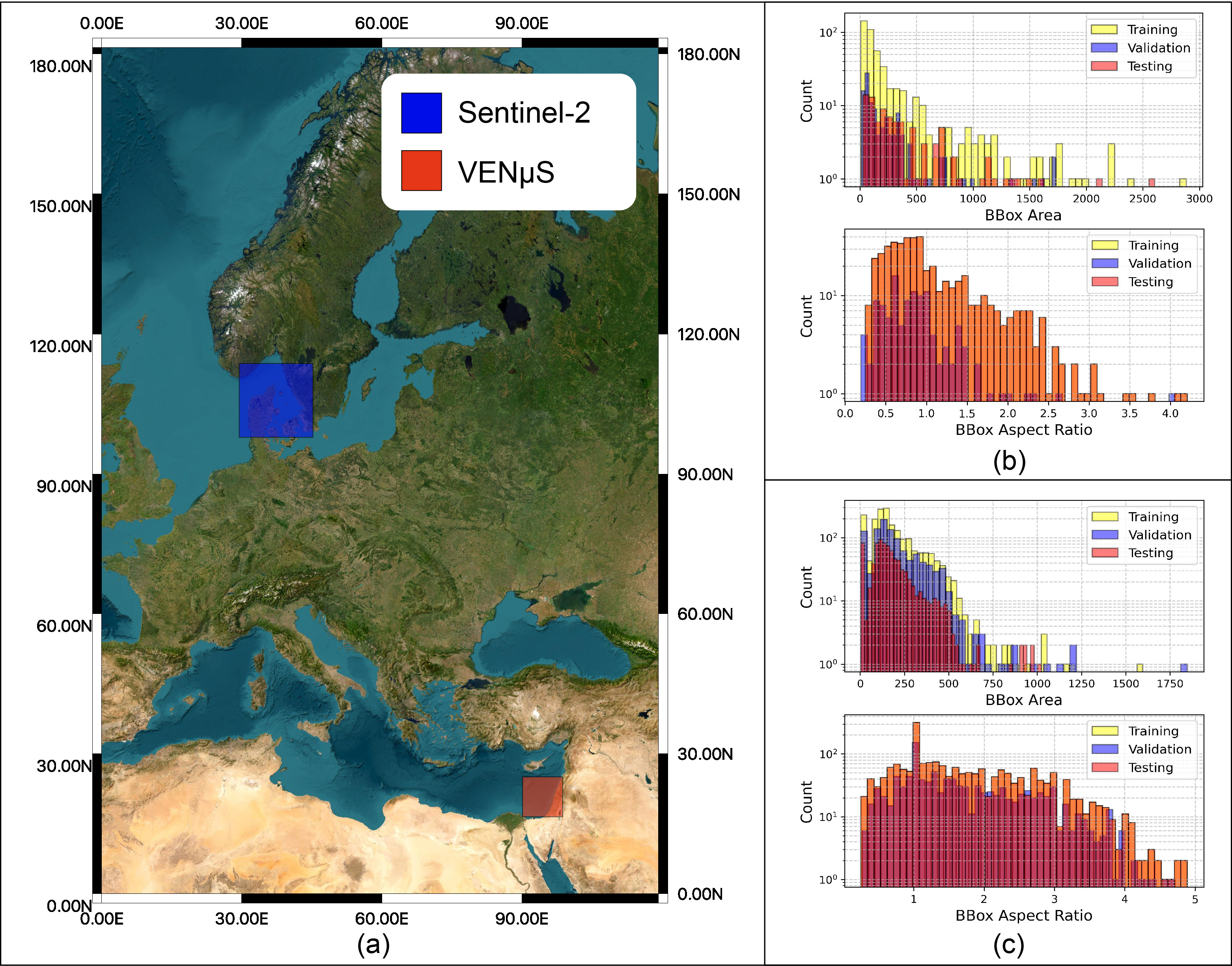}
        \caption{Graphical representation of the geographical coverage by the \gls{S-2} and VENµS datasets (a). The histograms illustrate the distribution of bounding boxes area and aspect ratio, offering a clear view of the training, validation, and test datasets for \gls{S-2} (b) and VENµS (c) missions.}
        \label{fig:datasets}
    \end{figure*}

    The data acquisition and labeling strategy is structured according to the algorithmic procedure outlined by Meoni et al. \cite{meoni2024unlocking}. 
    Initially, external information relevant to the task, specifically \gls{AIS} records, is collected. This is followed by utilizing \gls{L1C} data for labeling, where the \gls{L1C} images facilitate the clear visual detection of vessels and the retrieval of their latitude and longitude coordinates. Subsequently, a robust image matching technique \cite{sarlin2020superglue} is employed to transpose the annotations from \gls{L1C} to \gls{L0}. 
    Even though residual errors may occur due to inaccuracies in \gls{L0}-\gls{L1C} matching or band-to-band coregistration residual errors.
        To address these labelling inconsistency, we propose a coarse-to-fine bounding box strategy: we initially label the vessels with a coarse approximation, followed by applying an automated bounding box fitting technique to refine the accuracy across spectral bands.
        Specifically, we optimize the bounding box fit to the vessel using information derived from four distinct masking techniques. These techniques include \textit{Otsu}, \textit{Li}, \textit{Isodata}, and \textit{Mean} thresholding methods, and are used to separate foreground from background pixels. 
        \newline
        Otsu's Method \cite{otsu1975threshold} exploits the intra-class variance to segment vessels assuming that the image contains two classes of pixels, it calculates the threshold that separates these classes such that the intra-class variance is minimized.
        Li's Method \cite{li1998iterative} is an iterative approach to refining the threshold. It starts with an initial guess and iteratively updates the threshold based on a relationship that adjusts for differences between the current threshold and a function of it.
        The Isodata Method \cite{ridler1978picture}, also known as the Riddler-Calvar method, is another iterative scheme that determines the threshold by calculating the means of pixel intensities that fall below and above the current threshold value. This process continues until the threshold converges to a stable value. Finally, the 
        Mean Thresholding \cite{glasbey1993analysis} is a straightforward method where the threshold is simply set as the average intensity of all pixels in the image.
        The equations for each of these methods are presented below:
        \begin{equation}
        \begin{aligned}
            \textit{Otsu}:\quad & t = \underset{t}{\mathrm{argmin}} \left[\omega_{\text{bg}}(t) \sigma_{\text{bg}}^2(t) + \omega_{\text{fg}}(t) \sigma_{\text{fg}}^2(t)\right]; \\
            \textit{Li}:\quad & t_{i+1} = t_i + \frac{t_i - \text{Li}(t_i)}{1 + \text{Li}(t_i)}; \\
            \textit{Isodata}:\quad & t = \frac{m_L(t) + m_H(t)}{2}; \\
            \textit{Mean}:\quad & t = \frac{1}{N} \sum_{i=1}^{N} x_i
        \end{aligned}
        \end{equation}
        Notably, the terms used in the equations are defined as follows: \( \omega_{\text{bg}}(t) \) and \( \omega_{\text{fg}}(t) \) represent the probabilities of the background and foreground classes, respectively. The \( \sigma_{\text{bg}}^2(t) \) and \( \sigma_{\text{fg}}^2(t) \) denote the variances of the background and foreground classes. The terms \( m_L(t) \) and \( m_H(t) \) are the means of the lower and higher intensity classes, respectively. Finally, \( N \) is the total number of pixels, and \( x_i \) represents the intensity of the \( i \)-th pixel.
        \newline
        After thresholding, we then retain the pixels that are present in at least two of the segmentation maps (see Figure \ref{fig:thresholding}). 
        In the end, we iterate this step for each spectral band, producing an accurate vessel annotation within the \gls{L0} data for each of the spectral band under analysis. 
        \newline
    \newline
    In this manner, the two novel datasets have been compiled from scratch  using VENµS and Sentinel-2 \gls{MSI}s, each covering distinct geographic regions, as illustrated in Figure \ref{fig:datasets}(a).
    \newline
    These datasets cope to two specific tasks, namely detection and classification, obtained thanks to the AIS ancillary data,  consisting of information like the \gls{MMSI}, vessel type, dimensions (length and width), status, speed, position (longitude, latitude) and timestamp for each available vessel.
    It is important to acknowledge that AIS time intervals do not provide real-time ship data, due to the storage data policy of AIS vendors. Thus, in certain navigation scenarios, such as in narrow waterways, regions with high ship exchange flow density, or areas with extremely low navigation activity, AIS may become unreliable or invalid.
    \newline
    The next following paragraphs present the distinct characteristics of the VENµS and Sentinel-2 raw datasets, describing their mutual differences and highlighting their potential for scientific research. 

        \subsubsection{Sentinel-2 Dataset (VDS2Raw)}
        \label{sec:vds2raw}
       Building upon our previous study \cite{del2023first}, where we introduced the first raw multispectral dataset (VDS2Raw dataset), we employed the methodology outlined in our earlier work on THRawS \cite{meoni2023thraws}. In the study, we identified a reference dataset comprising Sentinel-2 \gls{L1C} products containing vessels. We utilized the dataset by Ruiloba et al. \cite{rosa_ruiloba_2020_3923841}, which includes Sentinel-2 \gls{L1C} tiles from 2019, acquired in Danish coastal areas. 
        To compile the dataset, we used a polygon surrounding the \gls{ROI} that included all ships for each of the eight acquisition days. We applied a sufficient margin to ensure that all the relevant bands were downloaded. This process led to the download of 390 \gls{L0} granules, which we then decompressed to obtain raw processable data.
        For labelling the dataset, we considered bands $B_i$ ($i=2,3,4$), using band $B_2$ serving as the reference for manual annotation. We created bounding boxes around each ship manually. To facilitate manual labelling, we used the coarse spatial coregistration technique described in our previous work \cite{meoni2023thraws}. We filled missing elements caused by the coregistration procedures using pixels from adjacent granules when possible \cite{meoni2023thraws}. In cases where this was not possible, we cropped the areas with missing pixels, resulting in granules with different pixel areas.
        After labelling, we selected 166 raw granules from the initial 390 downloaded granules. We split these into training (105), validation (27), and test (34) sets, with the number of annotations being 483, 119, and 93, respectively. The average granule size in the dataset is approximately estimated as \(2588.9 \, \text{px} \times 1669.4 \, \text{px}\), while the mean bounding boxes size is \(13.59 \, \text{px} \times 15.67 \, \text{px}\).
        As stated above, this process ensured that we had a comprehensive dataset, although it introduced variability in granule sizes due to the coregistration adjustments. 
        Further details regarding the creation of the initial version of VDS2Raw are available in \cite{del2023first}.
        \newline
        Compared to the preliminary version, we have implemented several enhancements to increase its utility for the scientific community. Here, we introduce VDS2Raw (v2), which will henceforth be referred to as VDS2Raw throughout this paper. First, we have expanded the dataset by incorporating additional ancillary \gls{AIS} records. The \gls{AIS} data come delivered free by the Danish government\footnote{https://web.ais.dk/aisdata/}.
        This \gls{AIS} source is particularly valuable due to its per-second sampling accuracy, which significantly enhances temporal resolution. 
        Second, we have included the $B_8$ band, providing an additional spectral band at 10m spatial resolution. This enhancement allows for more detailed and accurate analysis of the multispectral imagery, improving object detection and classification capabilities. We offer a more comprehensive and precise dataset by incorporating these improvements, facilitating more robust scientific research and applications.
        The vessel width and length distribution as the vessel classes are reported in Figure \ref{fig:sentivenus} (a). How it is possible to observe, the majority of vessels have a width of less than 10m with length of roughly 25m. The predominant ship classes are \textit{Cargo}, \textit{Tanker}, and \textit{Fishing} type, as shown in Figure \ref{fig:sentivenus} (c). 

        \subsubsection{VENµS Dataset (VDVRaw)}
\begin{figure*}[b]
            \centering
            \includegraphics[clip, trim=0 0cm 0 0, width=\linewidth]{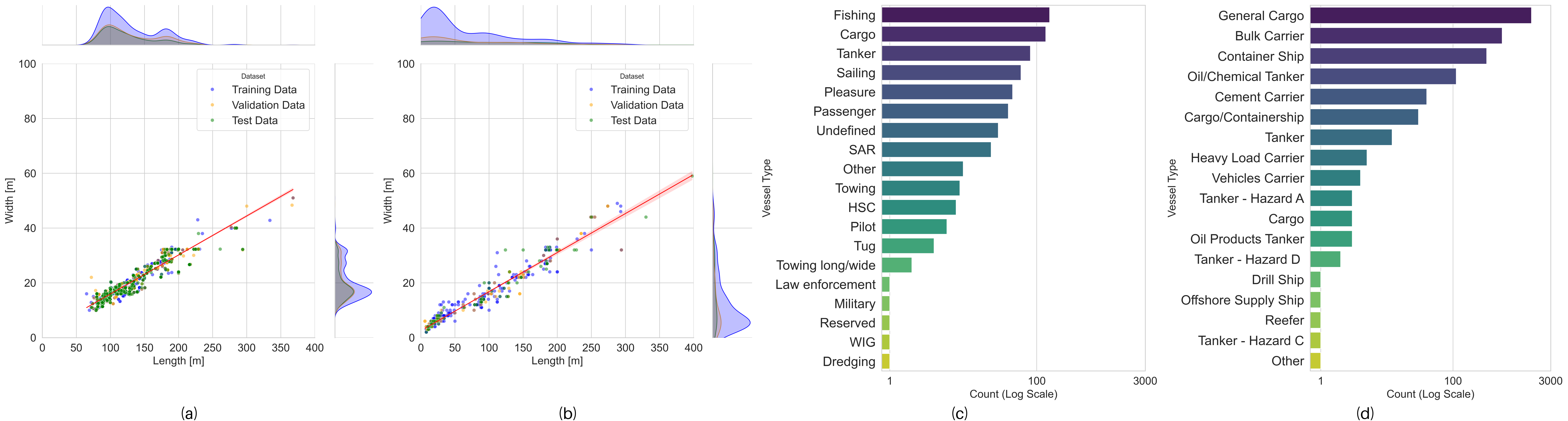}
            \caption{The AIS characteristics of the raw vessel datasets analyzed through various statistical visualizations. Histograms and kernel density estimates of length and width distributions are presented across training, validation, and test subsets for both the VDS2Raw (a) and VDVRaw (b) datasets. Additionally, bar plots (c) and (d) illustrate the frequency distribution of each vessel category within the VDS2Raw and VDVRaw datasets, respectively.
            }
            \label{fig:sentivenus}
        \end{figure*}

        The Vessel Detection from VENµS Raw (VDVRaw) is created using the \gls{ROI} encompassing the open sea adjacent to the port of Ashdod in Israel (Figure \ref{fig:datasets} (a)). The VENµS MSI captured imagery of this ROI approximately every two days at around 8:30 UTC during the initial phase of its scientific mission. AIS data for vessels within the ROI was sourced from a maritime analytics provider\footnote{MarineTraffic: www.marinetraffic.com}. However, the dataset presents a notable limitation: it only provides one AIS record per vessel per day within the ROI, typically recorded around midnight UTC. Consequently, information regarding a vessel's position at the precise moment the spacecraft sensed the ROI is unavailable.
        Nonetheless, it is common for vessels to remain stationary for extended periods, often several days, as they await authorization from port authorities to enter the port. This standard procedure enables port authorities to schedule maritime traffic efficiently based on the characteristics of the ships and prevailing conditions \cite{Ma2023}.
        \newline
        The database compiled for this study includes 3,827 vessels with bounding box labels in 282 multispectral images captured during the VENµS first mission phase. Among these vessels, 2,733 (71.4\%) have corresponding AIS records in the MarineTraffic database. The \gls{AIS} database categorizes these vessels into 18 different types. The Figure \ref{fig:sentivenus} (b) provides statistics on the length and width of vessels in the training validation and test dataset. Figure \ref{fig:sentivenus} (d) instead shows a bar plot counting each category reported in the AIS database. 
        As shown, the categories are quite unbalanced toward the \textit{General Cargo} class, with consequently more vessel dimension towards the length 100m and 15m of width, for all the datasets.   
        \newline 
        The database further incorporates flags ($F_l$) to indicate several vessel characteristics: presence of visible wakes ($F_l$=1), cloud cover ($F_l$=3), location on the image border ($F_l$=2), and proximity to other vessels ($F_l$=7). These flags can be combined to provide richer information about each vessel.
        The distribution of vessel counts by flag is as follows: 3,264 vessels (85.3\%) had no specific flag, 457 vessels (11.9\%) had visible wakes, 21 vessels (0.5\%) were located on the image border, 74 vessels (1.9\%) were under cloud cover, 4 vessels (0.1\%) were in proximity to other vessels, 1 vessel (0.0\%) had both cloud and wake flags, and 6 vessels (0.2\%) had both proximity and wake flags. 
        The final product of this process is a comprehensive dataset with AIS and ancillary information appended.
        
\subsubsection{Dataset Comparison}
As stated above, the two datasets encompass distinct geographical regions and span discrete time periods. As the Figure \ref{fig:sentivenus} displays, the vessel type distribution is quite dissimilar. In addition, despite the similar spectral responses of the bands under analysis (Fig. \ref{fig:VENuS_S2_bands}), their overall characteristics differ significantly. 
\newline
Although the predominant vessel classes in the VDS2Raw (v2) dataset remain \textit{Tanker} and \textit{Cargo}, this dataset encompasses a broader range of vessel types, including those that typically operate at higher speeds, such as \textit{Sailing}, \textit{SAR}, and \textit{Fishing} vessels. This variation is primarily attributed to the distinct \gls{AIS} records type utilized.
\newline
VENµS boasts superior spatial resolution compared to the \gls{MSI} of Sentinel-2, although it exhibits increased noise levels and artifacts in the acquired images. Additionally, compared to \gls{S-2}, the \gls{AIS} used for VDVRaw has been gathered with minor timing accuracy (once a day), further contributing to highlight the differences between the two. 
The purpose of this study is not to emphasize differences in model performance between the two datasets, but rather to highlight the similarities in how the models behave when applied to different types of raw data across distinct tasks. In summary, the goal of the present manuscript is to offer a representative analysis of model behavior when handling various kinds of raw data.
The differences between these two aforementioned data are summarised in the Table \ref{tab:compare_dataset}.

\begin{table*}[!b]
\footnotesize
\centering
\caption{Comparative table showing the differences between the VDS2Raw and VDVRaw datasets.}\label{tab:compare_dataset}
\begin{tabular}{@{} p{3cm}p{6cm}p{7cm} @{}}
\toprule
\textbf{Attribute} & \textbf{Sentinel-2} & \textbf{VENµS} \\ \midrule
Dataset Name & VDS2Raw (Official) & VDVRaw \\
Source & Raw granules from Sentinel-2 products & Images from VENµS satellite sensors \\
Spatial Resolution & 10m & 5.3m at Nadir \\
Spectral Bands & $B_{2}$, $B_{3}$, $B_{4}$, $B_{8}$ & 12, spanning visible to NIR \\
Reference Dataset & Ruiloba et al.\cite{rosa_ruiloba_2020_3923841} & From scratch \\
Area of Focus & Danish Fjords & Port of Ashdod, Israel \\
Time Period & 2019 & 2018 to end of October 2020 \\
Labeling & Manual, Polygons & Semi-Manual, Bounding Boxes \\
AIS Included & Yes & Yes \\
Sample Size (Avg) & 2588.9 px $\times$ 1669.4 px & 2798 px $\times$ 1929 px \\
BBox Size (Avg) & 13.59 px $\times$ 15.67 px & 14.32 px $\times$ 17.80 \\
Reference Band & $B_2$ & $B_5$ \\
Coregistration Technique & Coarse Spatial (Meoni et al. (2023)) & Affine Transform (SIFT) \\ \midrule
N. Samples & 166 (105 Training, 27 Validation, 34 Testing) & 284 (140 Training, 85 Validation, 57 Testing) \\
N. Annotations & 695 (483 Training, 119 Validation, 93 Testing) & 3.827 (1921 Training, 1061 Validation, 845 Testing) \\ \bottomrule
\end{tabular}
\end{table*}

    \subsubsection{AIS Records}
AIS records are essential for tracking maritime vessels, but they present several challenges. Firstly, data providers typically sample AIS messages at intervals of several minutes, leading to gaps in continuous tracking and historical \gls{AIS} records are often maintained only with a low frequency, further limiting their effectiveness for retrospective analysis. Secondly, raw multispectral imagery used for georeferencing can be imprecise, further complicating the task of accurately locating vessels. Thirdly, AIS messages themselves can be corrupted, falsified, or contain erroneous information as highlighted by \cite{balduzzi2014security} and \cite{ray2016methodology}. These issues collectively contribute to significant difficulties in matching labelled ship locations with those indicated by AIS data.
Given that the \gls{AIS} data in the two datasets differs significantly in temporal sampling, two distinct strategies were employed to address the matching problem.
\newline
For the VDS2Raw dataset, which features a high sampling rate, a structured multi-step approach was implemented. 
In the first step, metadata is extracted to facilitate temporal and geographical filtering of the \gls{AIS} data. This step retains the useful \gls{AIS} records, thereby focusing only on the most relevant data points.
In the second step, a cost matrix is constructed based on a weighted combination of perpendicular and Euclidean distances between the \gls{AIS} points and the bounding box centers. The weights of this matrix have been further adjusted to account the navigational status of the vessels, giving more importance to the vessel in movement. Then, an Hungarian algorithm \cite{kuhn1955hungarian} is systematically applied to optimally assign \gls{MMSI}s to the bounding box centers for each image granule.
Notably, to ensure consistency across all image granules, each \gls{MMSI} is uniquely assigned to a single bounding box. The assignment results have been stored and visually inspected to further verify the correctness of the matches.
%
%
%
%
%
\newline
It is important to state that, during the association process, we encountered instances where a single \gls{MMSI} was a candidate for different bounding boxes in near, but different, granules. Given also the complexities of handling densely trafficked areas, we deliberately avoided re-matching the unmatched bounding boxes with the globally-stored unmatched \gls{MMSI}s. Instead, we preferred to focus on determining the best association between a unique \gls{MMSI} and multiple bounding box candidates within the current set of processed granules. This decision is motivated by a lack of absolute trust in the \gls{AIS} data due to the aforementioned issues.
%
%
\newline
For the VDVRaw dataset, featuring only one daily record, the \gls{AIS} data was associated with the vessels based on a spatial and temporal criteria. 
Spatially, \gls{AIS} data points were selected within a 300-meter radius of the vessel’s center in the L1C image, ensuring spatial alignment. 
Temporally, to address the limitation of \gls{AIS} data providing only one daily record around midnight \gls{UTC}, a temporal window was applied, including records from the day the L1C image was captured, as well as the day before and after. The broader range improved the chances of accurately matching the \gls{AIS} record to the vessel’s position in the image.
This dual approach of combining spatial proximity with a flexible temporal range ensured that the vessel's position closely matched the corresponding \gls{AIS} records. This simple but effective approach was adopted because many vessels in the database remained stationary for extended periods. 
Nonetheless, any discrepancies were further resolved through visual inspection to ensure accurate matching.
\newline
%
%
%
%
In the end, to enhance the accuracy and generalizability of the \gls{ML} models, the issue of data imbalance is systematically addressed. This step is crucial, as both datasets exhibit substantial disparities in the distribution of vessel classes, as illustrated in Figure \ref{fig:sentivenus}(c) and Figure \ref{fig:sentivenus}(d).
First of all, a coarse-grained type classification has been chosen to perform. By focusing on these common and most promiscuous classes, we aim to create a more balanced and representative dataset for the classification task.
In the VDS2Raw dataset, we selected the vessel classes \textit{Cargo}, \textit{Fishing}, \textit{Sailing}, and \textit{Pleasure}, combining the last two (S\&P) due to their similar characteristics and to increase the sample size. For the VDVRaw dataset, we deliberately focused on the cargo-specific classes \textit{Bulk Carrier} and \textit{Container Ship}, excluding \textit{General Cargo} as its broader definition would introduce ambiguity and complicate classification. A comprehensive overview of two classification datasets is reported in Table \ref{tab:compare_dataset_class}.
\begin{table*}[!h]
\footnotesize
\centering
\caption{Comparative table summarizing the composition of the two classification datasets. For Sentinel-2, the classes \textit{Sailing} and \textit{Pleasure} have been combined (S\&P) due to their similar characteristics in order to increase the sample size; for VENµS, we deliberately excluded the \textit{General Cargo} class in order to focus on the cargo-specific classes.}
\label{tab:compare_dataset_class}
\begin{tabularx}{\linewidth}{@{}l XXX| XX@{}}
\cmidrule[\heavyrulewidth](lr){1-6} 

\multicolumn{1}{l}{\textbf{Attribute}} & \multicolumn{3}{l}
{\hspace{3cm}\textbf{Sentinel-2}} & \multicolumn{2}{l}{\hspace{2cm} \textbf{VENµS}} \\ 
\cmidrule(lr){1-6} 
 \hspace{0.4pt} Ship-types & \textit{Cargo} & \textit{S\&P} &  \textit{Fishing} & \textit{Bulk Carrier} & \textit{Container Ship}  \\ \cmidrule(lr){1-6}  
 \hspace{0.4pt} Train & 58 & 58 & 58 &  286 & 286 \\
 \hspace{0.4pt} Validation & 6 & 6 & 6 &  32 & 32 \\
 \hspace{0.4pt} Test & 42 & 35 & 59 &  325 & 80  \\ \cmidrule(lr){1-6} 

 \hspace{0.4pt} N. Samples (Tot.) & 106 & 99 & 123 &  643 & 398  \\ 
\cmidrule[\heavyrulewidth](lr){1-6} 
\end{tabularx}
\end{table*}

\noindent
To further address data imbalance and enhance model robustness, a combination of data augmentation techniques is employed, including random horizontal and vertical flips, random rotations up to forty degrees, and random perspective transformations. Additionally, a random elastic transform is applied to simulate realistic distortions. 
A toroidal shift, which shifts the image in a wrap-around manner resembling the shape of a torus, is also utilized. Random shifts within a specified range are applied both vertically and horizontally to introduce further variability into the augmented images. 

Careful consideration is given to preserving the metadata of the original images, conveniently embedding them along with the \gls{AIS} information directly within the COCO annotation file \cite{lin2014microsoft}. Specifically, we insert the \gls{MMSI}, information about vessel's type and route inside the annotation attribute of each vessel annotated, naming this extended format as AISCOCO.

%% file: sections/03_method.tex
\section{Processing Chain}\label{sec:method}

With the aim of improving \gls{MSA} by satellite technologies, past research focused on extracting information from SAR data \cite{eldhuset1996automatic,tello2006automatic,zhang2019ship} and multispectral data \cite{zhenwei_shi_ship_2014,mayra4827287mapping}. Concerning the latter, without loss of generality, we can cluster the methods into a) Threshold Segmentation Methods \cite{burgess1993automatic,corbane2008using,heiselberg_direct_2016}, b) Statistical Methods \cite{proia2009characterization,yang_ship_2014,corbane2010complete,10642217}, c) Feature-Based Methods \cite{zhenwei_shi_ship_2014,dalal2005histograms,guo2012novel}, and d) Deep-Learning Methods \cite{ciocarlan_ship_2021,mayra4827287mapping}.
\newline
The \gls{DL}-based techniques offer the advantage of not requiring manual feature extraction or threshold tuning, unlike threshold-based methods that rely on predefined value thresholds or statistical and feature-based methods that depend on specific models or descriptors. Their hierarchical representations within large datasets allow for greater generalization across varying imaging conditions and challenging environments \cite{ciocarlan_ship_2021,mayra4827287mapping}, as the abundant noise and artifacts present in raw data. 
\newline
The task of detection is to accurately identify ship location through bounding boxes. Differently, the primary objective for classification is to categorize the detected vessel correctly.
With the term identification, we intend the cascaded approach of detection and classification of each detected vessel.
It must be pointed out that an object detection framework enables the concurrent end-to-end detection and classification of instances, while in the current study, we chose to analyze these two tasks separately. 
By adopting this approach, we aim to disentangle the losses and the learned gradients, thereby optimizing each problem individually to achieve a better accuracy. This strategy facilitates a more effective understanding of each band's information across various tasks. 
\newline
In summary, the processing chain follows three essential steps:
\begin{enumerate}
    \item \textbf{Band-to-Band Registration}: Co-registration is performed to ensure accurate per-pixel alignment across bands. Processing efficiency is improved by activating co-registration only when multiple bands are demanded, thereby reducing computational overhead.

    \item \textbf{Detection}: DL models detect and locate vessels within the imagery, generating bounding boxes that focus subsequent analysis on relevant areas.

    \item \textbf{Classification}: Detected vessels are categorized into types (e.g., cargo ships, fishing boats) using a separate DL model, optimizing accuracy by isolating this task.

\end{enumerate}

\subsection{Band-to-Band Registration}
Band-to-Band registration is crucial in remote sensing image analysis, particularly when dealing with multispectral images. One of the main challenges in automatic alignment arises from significant non-linear intensity variations due to radiometric differences \cite{ye2014local}. This problem is especially severe in raw multispectral bands, where the lack of calibration intensifies these differences, making the registration process more complex. Ensuring accurate registration is essential as it directly impacts the performance of subsequent model inference and the overall effectiveness of remote sensing applications.
In the context of satellite operations, energy efficiency is of utmost importance. 
\newline
The work \cite{de2024improving} groups the registration methods in global and local.
For \gls{S-2}, we successfully adopted a statistically-based approach to conserve energy onboard, as described in our previous paper \cite{meoni2024unlocking}. This method falls in the global approaches as it provides with just two band offsets, for along-track and cross-track directions. Besides, it is specifically designed to optimize the registration process while minimizing energy consumption, thereby ensuring efficient and reliable satellite operations.
Concerning VENµS imagery, this approach was leading to considerable registration error ($>5px$) owing to the optical distortion at the borders of the stripmap acquistion. 
Therefore, for VENµS data we opted for a fine co-registration method employing SIFT \cite{lowe2004distinctive} keypoints: after identifying tie points, the bands are aligned using an non-linear transformation.

\subsection{Detection} 
For our benchmark, we select as vessel detector the model that in our previous study \cite{del2023first} was outperforming the others, i.e., VarifocalNet (VFNet) \cite{zhang2021varifocalnet}. 
We argue that the reason lies in its highly engineered head layers with: a) an tailored loss function \cite{zhang2021varifocalnet} and b) an anchor-free mechanism which directly classify and refine proposals without generating prior anchor boxes.
This dense object detector was born from the combination of the works of FCOS \cite{tian2022fully} and ATSS \cite{zhang2020bridging}. 
The model has been further made more lightweight trough the adoption of a residual convolutional neural network (ResNet-18) as backbone encoder for feature extraction. The multi-scale characteristics of vessels are addressed trough the usage of a Feature Pyramid Network (FPN) \cite{lin2017feature} module. This module, known as neck, merges the features of different levels with a bottom-up and a top-down pathway structure. Finally, the head layers, responsible for the bounding box regression and classification, are particularly convenient for small object detection thanks to the star-shaped deformable convolution (star dconv) \cite{zhang2021varifocalnet} applied for bounding box refinement, which inherently captures background sea information. 
%
%
\newline
Let $C_i$ represent the \textit{i-th} convolutional module where $i$ ranges from $1$ to $5$, the encoder is represented trough these five convolutional modules. Noteworthy, the stride of each convolutional module is two, meaning that spatial dimension is halved after each $C_i$ forming the feature pyramid top-down pathway. 
Each feature map is then convolved to \textit{d=256} channels by 1$\times$1 convolution, making the $M_i$ feature maps. The parameter $d$ was set according to the original paper \cite{zhang2021varifocalnet}.
For $i$ spanning from $5$ to $3$, the feature map $M_i$ is upsampled and added to the $M_{(i-1)}$ following bottom-up pathway. A 3$\times$3 convolution is further applied to the feature maps $M_2$, $M_3$, $M_4$. The feature maps resulting from this process are termed $P_i$.
VFNet head layers are subsequently applied to each feature map $P_i$. Each head layer is composed of two sub-networks, one for bounding box regression and one for classification. 
These networks show similar characteristics, employing three 3$\times$3 convolutions followed by ReLU activation. The regression network is highly engineered adopting a bounding box refinement scheme where the initial bounding box coordinates are improved thanks to the information sampled at the nine points identified by the star dconv \cite{zhang2021varifocalnet}. Notably, this sampled information is also passed to the classifier branch to improve the classification performance. The number of channels in output equals the number of classes considered for the problem.
In the end, \gls{NMS} is applied to all suppress all the predictions from the diverse heads to remove spurious detections. This component is necessary due to the fact that the predictions are not one-to-one assigned but rather a series of predictions produced by the network.

\subsection{Classification}
After vessel localization, category classification is performed using the ResNet-18 feature extractor as in the detection encoder. ResNet-18, conceptualized as an exponential ensemble of shallow networks \cite{veit2016residual}, mitigates vanishing gradients via ``skip connections'' that enable direct gradient propagation across layers, ensuring efficient backpropagation. The 18-layers architecture balances depth and computational efficiency, making it optimal for high-accuracy, low-latency tasks.
Unlike detection, we leverage only the final layer's feature map, containing the most abstracted representations. A classification head is appended to these high-level features to categorize vessel types effectively.

%% file: sections/04_results.tex
\section{Experimental Results}\label{sec:results}

This section rigorously evaluates the model's performance on the custom-developed datasets across single and multi-band inputs. The following Table \ref{tab:model_evaluation} provides a comprehensive overview of the model evaluation and training methodologies used in this study. 
%

\begin{table}[h]
\caption{Summary of Model Evaluation, Optimization, and Training Techniques}
\label{tab:model_evaluation}
\centering 
\begin{tabularx}{\columnwidth}{p{4.5cm}X} 
    \toprule
    \textbf{Setting} & \textbf{Description} \\
    \midrule
    \textbf{Training Setup} & Linux-based system, NVIDIA A100 GPU (40GB), 16-core AMD EPYC-Rome processor at 2.8 GHz. \\
    \textbf{Hyperparameter Optimization} & Grid search optimizing learning rate, epochs, batch size, learning rate schedule \\
    \textbf{Early Stopping} & Patience of 30 epochs \\
    \textbf{Data Augmentation} & Rotations, flips, scaling, shearing to increase variability and model generalization (\textbf{detection}); Rotations, flips, perspective transformations, elastic deformations (\textbf{classification})\\
     
    \textbf{Input Size} & 2048 $\times$ 2048 (\textbf{detection}); 64 $\times$ 64 (\textbf{classification})\\
    \textbf{Optimizer}  & Stochastic Gradient Descent (momentum=0.9, weight decay=0.0001) (\textbf{detection}); AdamW (weight decay=0.01) (\textbf{classification}) \\
    \textbf{Learning Rate Scheduler} & Linear warm-up (10\% epochs), multi-step decay (factor 0.8 up to 50\% of epochs), cosine annealing (final phase) \\
    \bottomrule
\end{tabularx}
\end{table}

It highlights key aspects such as the performance evaluation on custom-developed, the computational setup, and the systematic approach to hyperparameter optimization. Additionally, it summarizes the data augmentation techniques and training strategies employed to enhance model generalization.
In Section \ref{sec: metrics}, we clarify the adoption of certain metrics for model evaluation. Sections \ref{sec:singleanalysis}-\ref{sec:multianalisi} aim to demonstrate the advantages and disadvantages of using a single or a combination of spectral bands.
By comparing the performance of single and multiple spectral bands, the study will elucidate the conditions under which synergistic exploitation is most beneficial.
This structured approach ensures a detailed evaluation of the spectral bands. Specifically, we tested the spectral bands from \gls{S-2} at the highest spatial resolution (10m), including individual bands and their combined usage. For VENµS, which has all the bands at the same spatial resolution (5m), the analysis encompasses bands $B_1$ through $B_{12}$, alongside their combinations.
\newline
It must be noted the performance over a particular band or set of bands may be influenced by the specific characteristics of the data set used in this study, raising questions about their generalizability to other contexts or datasets. We cope with this issue by analysing the behaviours of the bands in the two mentioned datasets. 

\subsection{Metrics}\label{sec: metrics}
Several considerations are critical in the context of vessel identification from raw multispectral data, as the lack of calibration, band-to-band registration, and geometric corrections can hamper model performance. 
Additionally, varying conditions, e.g. atmospheric, sea and water reflections, further complicate the tasks. 
Therefore, an effective evaluation must take into account these effects, selecting appropriate metrics for accurately assessing the model's ability to locate and categorize vessels in each band. 
\newline
The \gls{IoU} \cite{lin2014microsoft} is common among the detection metrics and measures the similarity between predicted and actual bounding boxes. Given two shapes $A$ and $B$, IoU is the ratio of their overlap area to their union area. It is used to classify true positives (TP) and false positives (FP) based on a specified threshold. Denoting FN as false negatives, from IoU the metrics of Precision (P) and Recall (R) and F-1 score ($F_1$) can be derived as:
\begin{equation}\label{eq:IOU}
\text{IoU} = \frac{|A \cap B|}{|A \cup B|}, \quad P = \frac{\text{TP}}{\text{TP} + \text{FP}}, \quad R = \frac{\text{TP}}{\text{TP} + \text{FN}}, \quad F_1 = 2 \cdot \frac{\text{P} \cdot \text{R}}{\text{P} + \text{R}}
\end{equation}
We note that high Recall is crucial to ensure that all vessels, including smaller ones, are detected, while high precision is essential to minimize the number of false alarms, ensuring proper resource allocation from marine agencies. Finally, the F1-score provides a balanced measure of the model's overall performance. These metrics are a common choice for object detection benchmarks.
\newline
Notwithstanding, these metrics face difficulties when the problem is shifted from generic object detection to small objects owing to the limited amount of spatial features \cite{chen2020survey} and significant variance of \gls{IoU} at lower scales \cite{jeune2023rethinking}.
The variance arises from the scale-invariant nature of \gls{IoU}: if the \gls{IoU} between predicted ($b_1$) and a ground truth ($b_2$) boxes is equal to $\text{IoU}(b_1, b_2) = u$, then scaling all coordinates of both bounding boxes by a factor $k$ results in the same \gls{IoU} value:  
$\text{IoU}(b_1, b_2) = \text{IoU}(kb_1, kb_2) = u$.  
\newline
In contrast, the ratio ($\varepsilon_{\text{loc}} / \omega$) between the localization error ($\varepsilon_{\text{loc}} = \|b_1 - b_2\|_1$) and the object size ($\omega$) increases as the objects become smaller. Hence, \gls{IoU} is particularly sensitive to spatial alignment and bounding box precision when dealing with small objects, as highlighted in the study of \cite{jeune2023rethinking}. 
This is particularly relevant since detection models lack of scale-invariant properties as they do not localize small and large objects with the same accuracy level. 
\newline
In conclusion, the scale-invariant property is advantageous for classical object detection but has problems when dealing with small object detection because it actually causes \gls{IoU} to be overly sensitive to small localization errors, affecting metrics evaluation \cite{jeune2023rethinking}. \gls{IoU} drops dramatically when the localization error increases for small objects, without diminishing the quality of the detection from a human perspective \cite{jeune2023rethinking}.
\newline
While past research \cite{rezatofighi2019generalized,zheng2020distance,he2021alpha,NWD,guo2023robust, tran2021updated} was conceived on formulating \gls{IoU}s tailored for the regression loss, few attention has been dedicated to small object detection and its evaluation.
Recently, \cite{jeune2023rethinking} reflected on the variance issue highlighting the importance of using a low-variance criterion for evaluation, particularly when dealing with detectors that, on average, produce well-localized bounding boxes. Indeed, an high-variance criterion can lead to inconsistent evaluations, where even adequately localized boxes may be incorrectly classified as false negatives due to random variations in performance. 
Addressing these challenges, the Scale-adaptive IoU (SIoU) \cite{jeune2023rethinking} has been proposed as a novel similarity metric in small object detection. 
To mitigate the high variance of \gls{IoU} at smaller scales, SIoU introduces a scaling factor $p$ that dynamically adapts according to the size of the objects, making the metric more lenient towards small shifts in smaller objects.
The mathematical formulation of SIoU is given by:
\begin{equation}
\text{SIoU}(b_1, b_2) = \left(\text{IoU}(b_1, b_2)\right)^p, \quad \text{where} \quad p = 1 - \gamma \exp\left(-\frac{\sqrt{w_1 h_1 + w_2 h_2}}{\sqrt{2}\kappa}\right)
\end{equation}
Here, $b_1$ and $b_2$ are the bounding boxes with dimensions $w_1, h_1$ and $w_2, h_2$, respectively. The parameter $\gamma$ controls the degree of scaling, particularly for smaller objects, while $\kappa$ influences the rate at which the scaling effect diminishes as object size increases. The exponential function ensures that the scaling factor $p$ decreases gradually based on the bounding boxes' sizes.
Since SIoU yields higher expected values than IoU for small objects \cite{jeune2023rethinking}, it compensates for greater localization errors, offering a more robust and perceptually consistent metric for object detection. In this study, we set the parameter \(\gamma = 0.5\), following the original work by \cite{jeune2023rethinking}, while \(\kappa = \sqrt{8}\) has been chose empirically.
\newline
Unlike object detection, classification deals solely with assigning a label to an entire tile, assuming that the object of interest is already localized. This difference means that classification models emphasize discriminative power among different classes without the additional burden of spatial accuracy.
When dealing with raw multispectral data, where spectral bands offer additional discriminative features, the challenge lies in effectively leveraging this rich information to distinguish between classes. Moreover, since the datasets available contains very few samples in comparable categories, the choice was to adopt for \gls{MCC} as the metric to evaluate and compare the training models, which already found numerous applications for assessing classification problems on unbalanced datasets \cite{Delgado2019WhyCK, Chicco2021TheBO}. Using the same nomenclature of Precision and Recall, the \gls{MCC} can be formulated as:
\begin{equation}
\text{MCC} = \frac{\text{TN} \cdot \text{TP}-\text{FP} \cdot \text{FN}}{\sqrt{\left(\text{TP}+\text{FP}\right)\left(\text{TP}+\text{FN}\right)\left(\text{TN}+\text{FN}\right)\left(\text{TN}+\text{FP}\right)}}
\end{equation}


\subsection{Single-Band Analysis}\label{sec:singleanalysis}
\subsubsection{VDVRaw}

\begin{figure}[htb]
    \centering
    \includegraphics[width=\linewidth,clip]{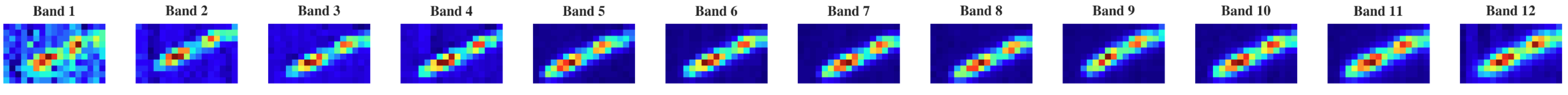}
    \includegraphics[width=\linewidth,clip]{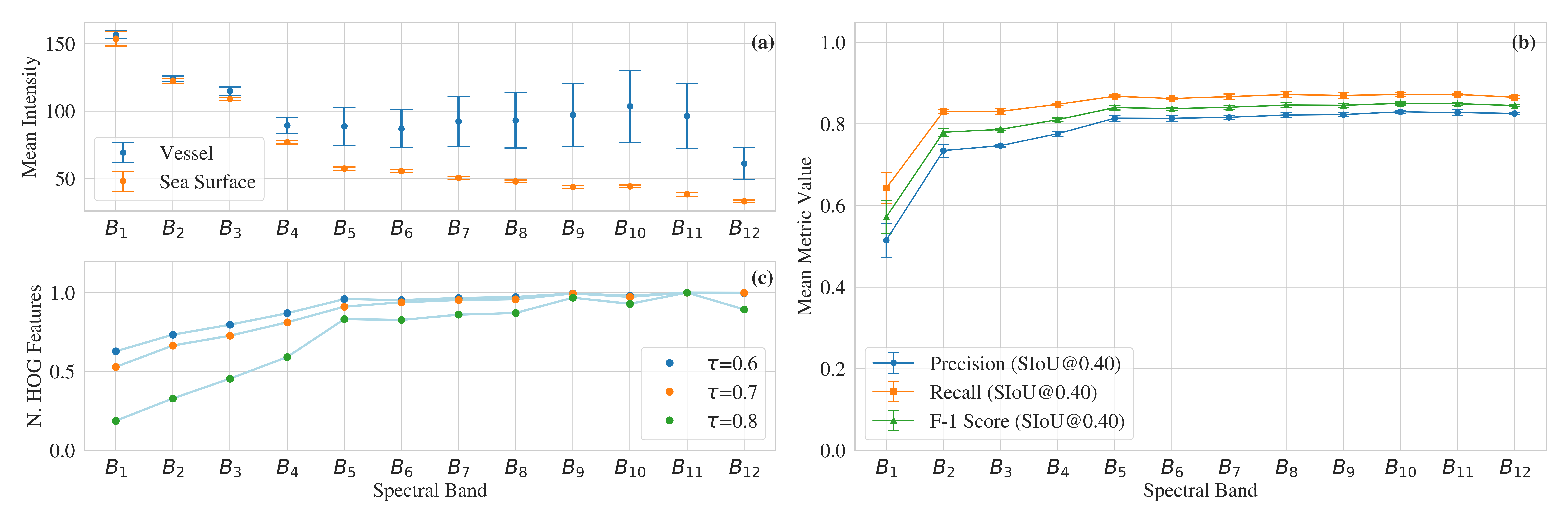}
    \caption{VENµS (VDVRaw) mean intensity (\gls{DN}) for vessel and sea clutter (a), mean detection metric values (b), and normalized number of \gls{HOG} features (c) in each spectral band under consideration. Fluctuation of the vessel in the various bands due to the residual co-registration errors.}
    \label{fig:figure9}
\end{figure}

Several factors significantly influence the overall detection accuracy of the studied model, including the spectral response of each band and its associated noise level.
Figure \ref{fig:figure9} illustrates these factors for each spectral band. The top panel displays a vessel (Cargo Type) while the bottom panel presents (a) the mean intensity values of vessels and sea surface, (b) the number of \gls{HOG} \cite{dalal2005histograms} features computed at various thresholds ($\tau$), and (c) the Precision, Recall, and $F_1$ scores evaluated at an SIoU threshold of 0.40.
The number of \gls{HOG} features is considered a measure of the information content within each spectral band. \gls{HOG} calculated with a $2 \times 2$ kernel provide insights into the amount of variability and, consequently, the information contained within the band. By applying different thresholds, we identified a growing pattern in information content from $B_1$ to $B_{12}$, as shown in Figure \ref{fig:figure9} (b). Furthermore, to account for the quality of the input information to our model, we indirectly measure the noise by calculating the average spectral response over the sea surface. We then compare it with respect to the average spectral response of vessels in each band, noting an interesting behaviour. 
Recalling that spectral bands are uncalibrated, the model's performance shows a direct correlation with the number of features present in the bounding boxes for each band. 
We measure this ``quantity of information'' by estimating the number of strong gradients, or changes, in adjacent $2 \times 2$ kernels. 
Specifically, we use the \gls{HOG} feature descriptor \cite{dalal2005histograms} with different thresholds.
The two kinds of information, quality (a vessel-to-sea ratio) and quantity (number of strong gradients) enable us to identify a correlation with the model results.
\newline
By looking at Figure \ref{fig:figure9}(a), it is possible to notice a descending trend in the average spectral response (expressed in \gls{DN}) of the sea surface moving from lower to higher wavelengths. A similar trend is present in the number of \gls{HOG} features, this time growing from lower to higher bands.

\begin{table}[htb]
\centering
\footnotesize
\caption{VENµS (VDVRaw) evaluation metric (MCC) computed for different single spectral bands for classification.}
\label{tab:metrics_s2_single}
\begin{tabularx}{\textwidth}{p{1.5cm}*{12}{>{\centering\arraybackslash}X}}
\toprule
\textbf{Spectral Bands} & \textbf{B$_{1}$} & \textbf{B$_{2}$} & \textbf{B$_3$} & \textbf{B$_4$} & \textbf{B$_{5}$} & \textbf{B$_{6}$} & \textbf{B$_{7}$} & \textbf{B$_8$} & \textbf{B$_{9}$} & \textbf{B$_{10}$} & \textbf{B$_{11}$} & \textbf{B$_{12}$} \\
\midrule
\textbf{Mean} & 0.33 & 0.48 & 0.50  & 0.52  & 0.57  & 0.58  & 0.57  & 0.58  & 0.56  & \textbf{0.62 } & 0.57  & 0.51  \\
\textbf{Std} & ±0.02 & ±0.05 & ±0.04 & ±0.05 & ±0.02 & ±0.03 & ±0.03 & ±0.04 & ±0.04 & \textbf{±0.02} & ±0.02 & ±0.02 \\
\bottomrule
\end{tabularx}
\end{table}

\begin{figure}[!h]
    \centering
    \includegraphics[width=\linewidth]{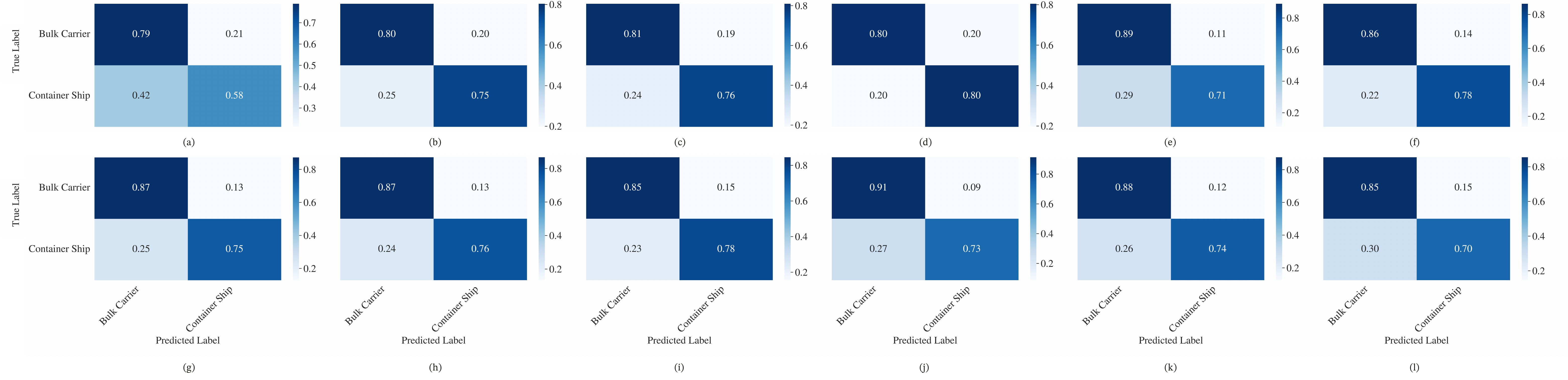}

    \caption{VENµS (VDVRaw) confusion matrices for each spectral band under consideration. The bands are labeled as follows: (a) B1, (b) B2, (c) B3, (d) B4, (e) B5, (f) B6, (g) B7, (h) B8, (i) B9, (j) B10, (k) B11, and (l) B12.}
    \label{fig:vdvraw_single_conf_matrices}
\end{figure}

The bands $B_1$ and $B_2$ are particularly sensitive to water clarity and depth variations, thus resulting much more affected by noise. Band $B_1$ shows the lowest performance with a mean Precision of 0.42, mean Recall of 0.54, and mean $F_1$ Score of 0.47. The bands $ \{ B_i \}_{i=2}^{5}$ show a progressive improvement in all metrics, peaking at band $B_5$ that achieves a mean Precision of 0.75 and a mean Recall of 0.82, resulting in a mean $F_1$ Score of 0.78.
The highest-performing bands are $ \{ B_i \}_{i=6}^{12} $, which are characterized by similar metric values. Among these, the bands $B_{10}$, $B_{11}$, and $B_{12}$ have the highest mean $F_1$ Scores (around 0.81), and provide the most stable performance across all metrics.
\newline
Classification results display a similar, even though less pronounced, trend. This is also reflected by the stable performance values displayed in the confusion matrices of Figure \ref{fig:vdvraw_single_conf_matrices}. Among the twelve spectral bands, band $B_10$ achieves the highest MCC value of 0.62 (±0.02), suggesting it provides the most discriminative power for the two-class problem at hand. Bands $B_6$, $B_8$, and $B_11$ also demonstrate relatively strong performance with MCC values of 0.57 to 0.58, though their variation remains within a narrow range, reflecting the stable yet modest differentiation capability across these bands.
\subsubsection{VDS2Raw}

\begin{figure}[!b]
        \centering
        \includegraphics[width=\linewidth]{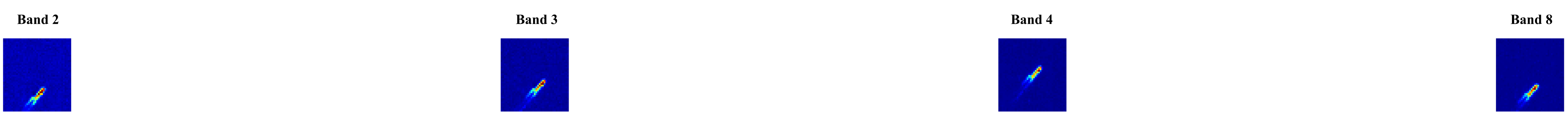}
        \includegraphics[width=\linewidth]{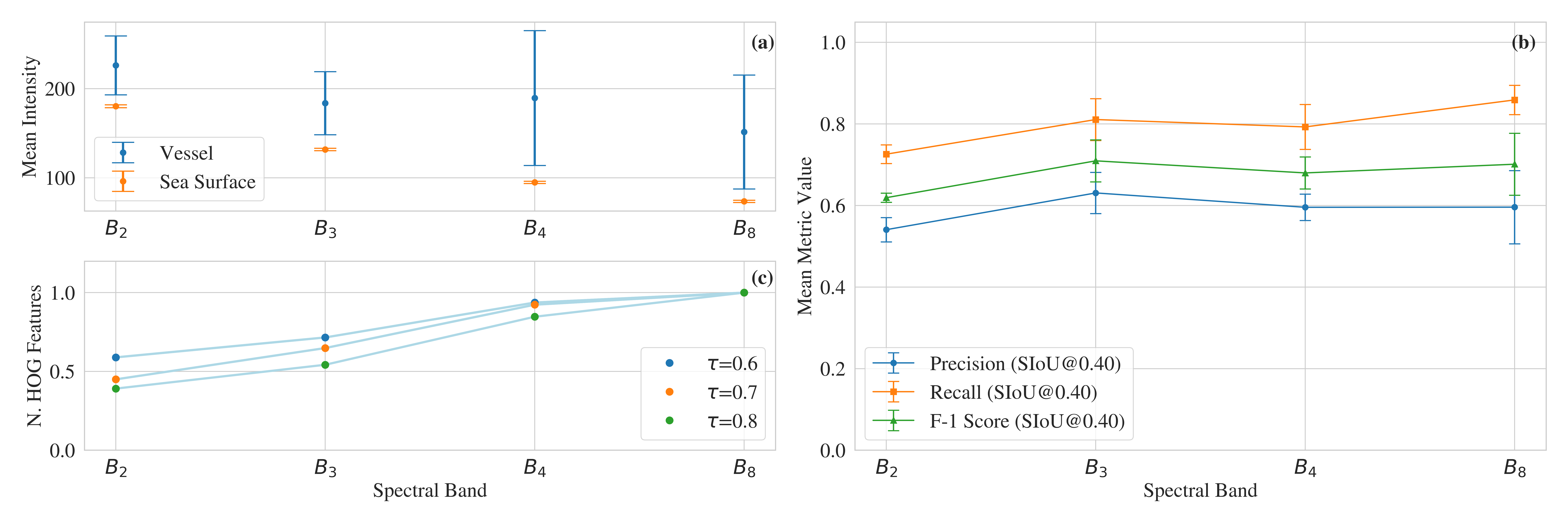}
        \caption{Sentinel-2 (VDS2Raw) mean intensity for vessel and sea clutter (a), mean detection metric values (b), and normalized number of \gls{HOG} features (c) in each spectral band under consideration. Fluctuation of the vessel in the various bands due to the residual co-registration errors.}
        \label{fig:vds2raw_single}
    \end{figure}

Consistently with VENµS, we assess the performance of the model on the several bands of \gls{S-2}. Figure \ref{fig:vds2raw_single} provides an overview of the key factors for each spectral band under consideration ($ B_i \quad \text{for} \quad i \in \{2, 3, 4, 8\} $). 
The top panel in Figure \ref{fig:vds2raw_single}(a) shows the mean intensity values for vessels and sea clutter, which serve as an indicator of the quality of the input data. Similar to our analysis for VENµS, the vessel-to-sea ratio is a critical measure for understanding the noise characteristics in each band.
Figure \ref{fig:vds2raw_single}(b) displays, like in the previous paragraph, the mean values of the detection metrics—Precision, Recall, and $F_1$ score—for each spectral band. It is evident that certain bands consistently outperform others, with bands closer to NIR demonstrating the highest values across all three metrics. This observation aligns with the trend observed for VENµS, where NIR bands with a higher number of strong gradient features and better vessel-to-sea ratios yielded superior detection performance.
The bottom panel of Figure \ref{fig:vds2raw_single}(c) shows again the normalized number of \gls{HOG} features calculated for each band at varying thresholds. As with VENµS, the number of \gls{HOG} features serves as a measure of the information content in each band. Also in this case, a consistent increase in the number of \gls{HOG} features is observed from increasing wavelengths. 
\newline
Once again, when examining both the vessel-to-sea ratio and the number of strong \gls{HOG} features, the duality information content---quality and quantity---enables us to draw a correlation with the model's detection metrics. Differently from the VENµS case, the improvements are less prominent with increasing wavelengths. In particular, the bands $B_3$ and $B_8$ scored higher Precision and Recall values, respectively. The results in terms of $F_1$ score lie consistently around 0.7 for the $B_3$, $B_4$ and $B_8$ bands.
\newline
In conclusion, consistent with the findings from VENµS, the Sentinel-2 analysis shows that bands lying in spectral regions close to 800$nm$ demonstrate remarkable results in Precision, Recall, and $F_1$ scores. This further corroborates the relationship between the number of strong \gls{HOG} features in bounding boxes for each band and the model's performance, underscoring the importance of both spectral response and noise levels in vessel detection from raw data.

\begin{table}[htb]
\centering
\footnotesize
\caption{Sentinel-2 (VDS2Raw) evaluation metric (MCC) computed for different single spectral bands for classification.}
\label{tab:metrics_s2_single}
\begin{tabularx}{\textwidth}{p{1.5cm}*{4}{>{\centering\arraybackslash}X}}
\toprule
\textbf{Spectral Bands} & \textbf{B$_{2}$} & \textbf{B$_3$} & \textbf{B$_4$} & \textbf{B$_8$} \\
\midrule
\textbf{Mean} & 0.31& \textbf{0.35 } & 0.33  & 0.33 \\
\textbf{Std} & ±0.05 & \textbf{±0.05} & ±0.10 & ±0.01 \\
\bottomrule
\end{tabularx}
\end{table}

\begin{figure}[!h]
        \centering
        \includegraphics[width=\linewidth]{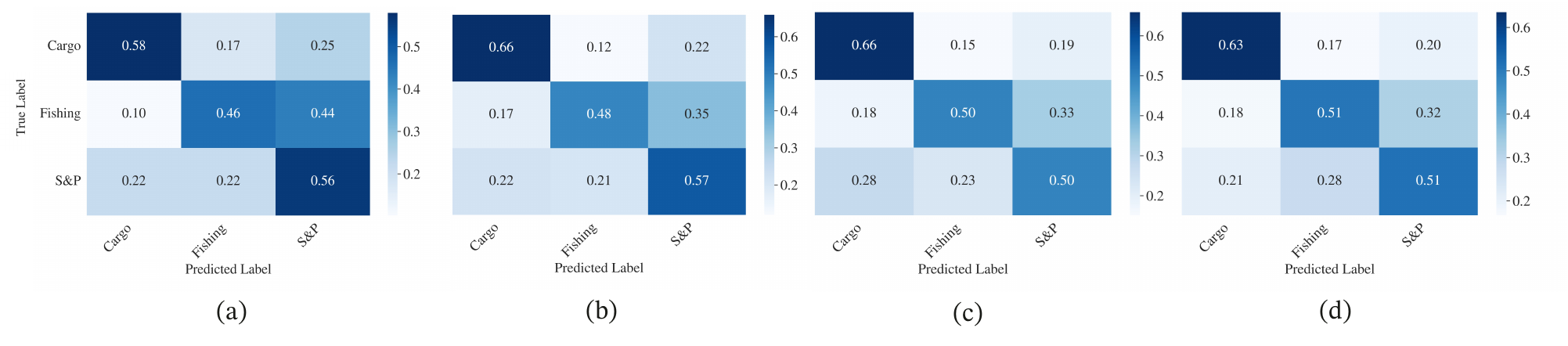}

        \caption{Sentinel-2 (VDS2Raw) confusion matrices in each spectral band under consideration. The bands are labeled as follows: (a) B2, (b) B3, (c) B4, and (d) B8.}
        \label{fig:vds2raw_single_conf_matrices}
    \end{figure}

%
Concerning classification scores, the Figure \ref{fig:vds2raw_single_conf_matrices} remarks how even for \gls{S-2} the results remained consistent across all spectral bands. 

\subsection{Multi-Band Results}
\label{sec:multianalisi}

\subsubsection{VDVRaw v2}
\begin{figure}[tb]
            \centering
            \includegraphics[width=0.7\linewidth]{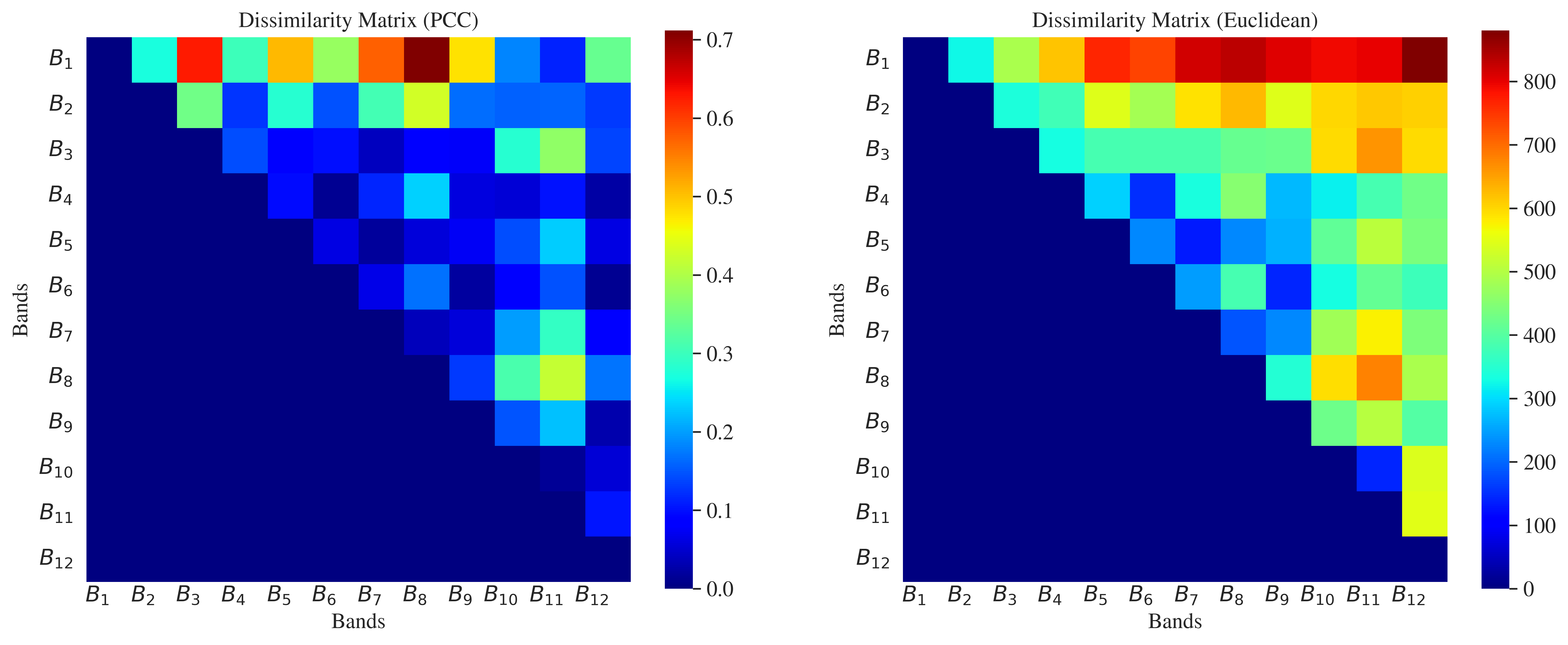}
            \caption{Dissimilarity matrices obtained over the VDVRaw dataset (bbox only): \gls{PCC} (left) and \gls{ED} (right).}
            \label{fig:dissimatrix}
        \end{figure}

        For VDVRaw, instead of evaluating all possible combinations from the 12 spectral bands of the VENµS MSI, we focused on selecting the most dissimilar bands to provide the model with diverse, complementary information. We used two metrics to quantify band dissimilarity: the \gls{PCC} and \gls{ED}. The \gls{PCC} measures linear relationships between bands, while the \gls{ED} assesses their geometric separation. Together, these metrics help identify band combinations that maximize informational diversity.
        The \gls{PCC} is defined as:
        \begin{equation}
            \rho_{B_iB_j} = \frac{\text{cov}(B_i, B_j)}{\sigma_{B_i} \sigma_{B_j}},
        \end{equation}
        where \(\rho_{B_iB_j}\) is the correlation coefficient between bands \(B_i\) and \(B_j\), \(\text{cov}(B_i, B_j)\) is their covariance, and \(\sigma_{B_i}\), \(\sigma_{B_j}\) are their standard deviations.
        The \gls{ED} between bands \(B_i\) and \(B_j\) is:
        \begin{equation}
            d(B_i, B_j) = \sqrt{\sum_{k=1}^{n} (B_{i,k} - B_{j,k})^2},
        \end{equation}
        where \(d(B_i, B_j)\) is the distance, and \(B_{i,k}\), \(B_{j,k}\) are the \(k\)-th elements of \(B_i\) and \(B_j\). 
        \newline
        The analysis focused only on the bounding boxes around vessel structures to ensure relevant regions were considered. Figure~\ref{fig:dissimatrix} shows the dissimilarity matrices obtained using \gls{PCC} and \gls{ED}, illustrating the dissimilarity between spectral bands to guide the selection of the most informative ones.
        \newline
        Based on the dissimilarities identified, we selected various sets of band combinations to optimize the model's performance. In doing so, we intentionally excluded lower-performing bands identified in previous analyses to focus on combinations that maximize the diversity of information provided to the model. This selective approach is designed to enhance the model's ability to differentiate and detect features more effectively by leveraging the most distinct and informative spectral bands.
        \newline
        For two-band combinations, three pairs emerged as particularly significant: $B_5$ (visible) with $B_{8}$ (visible), $B_5$ (visible) with $B_{10}$ (near-infrared) and $B_5$ (visible) with $B_{12}$ (near-infrared). These pairs were notable for their distinct spectral information, making them effective in distinguishing between various features within the data. The strength of these combinations lies in their simplicity and ease of implementation, which can be highly beneficial in applications requiring quick, preliminary analysis. However, the reliance on only two bands may limit the breadth of spectral variation captured, potentially overlooking subtler features that could be detected with additional bands.
        \newline
        For this reason, we identified two other sets of three-band combinations: $B_3$, $B_4$, and $B_7$ (all visible RGB) as well as $B_5$ (visible) with $B_{10}$ and $B_{11}$ (both NIR). The strength of these combinations is their ability to capture a wider range of spectral variation, thus offering more variegated insights into the structures of vessels under analysis. 
        Finally, for four-band combinations, two key sets were highlighted: $B_3$, $B_4$, and $B_7$ (all visible) with $B_{11}$ (near-infrared), as well as $B_3$, $B_5$, and $B_7$ (visible) with $B_{11}$ (near-infrared). These combinations are particularly valuable as they offer a comprehensive view of the spectral characteristics, enabling a deeper and more detailed understanding of the data. 
        \newline
        We evaluated the different band combinations using the same strategy applied to the individual bands, with the results presented in Table \ref{tab:multi_venus}. 
        \begin{table}[htb]
        \centering
        \footnotesize
        \caption{VENµS (VDVRaw) evaluation metrics computed for different spectral band combinations: Precision, Recall, $F_1$ score for detection, and MCC for classification.}
        \label{tab:metrics_combinations}
        \begin{tabularx}{\columnwidth}{p{3cm}p{3cm}p{3cm}p{3cm}|l}
        \toprule
        \textbf{Spectral Bands}             & \textbf{Precision}      & \textbf{Recall}         & \textbf{$F_1$}       & \textbf{MCC} \\ 
        \midrule
        \textbf{B$_{10}$}              & 0.82 (±0.01)            & 0.87 (±0.00)            & 0.85 (±0.00)            & \textbf{0.62 (±0.02})                         \\
        \textbf{B$_5$B$_8$}              & 0.80 (±0.01)            & 0.86 (±0.00)            & 0.83 (±0.00)            & 0.53 (±0.03)                              \\
        \textbf{B$_5$B$_{10}$}           & 0.81 (±0.01)            & 0.86 (±0.00)            & 0.84 (±0.00)            & 0.56 (±0.03)                               \\
        \textbf{B$_5$B$_{12}$}           & 0.81 (±0.01)            & 0.87 (±0.00)            & 0.84 (±0.01)            & 0.53 (±0.06)                               \\
        \textbf{B$_5$B$_{10}$B$_{11}$}   & 0.84 (±0.01)            & 0.86 (±0.01)            & 0.85 (±0.01)            & 0.57 (±0.00)                                \\
        \textbf{B$_3$B$_4$B$_7$}         & \textbf{0.85 (±0.00)}   & 0.87 (±0.00)            & \textbf{0.86 (±0.00)}   & 0.60 (±0.01)                          \\
        \textbf{B$_3$B$_4$B$_7$B$_{11}$} & 0.84 (±0.01)            & 0.87 (±0.00)            & \textbf{0.86 (±0.00)}   & 0.58 (±0.01)                                 \\
        \textbf{B$_3$B$_5$B$_7$B$_{11}$} & \textbf{0.85 (±0.00)}   & \textbf{0.88 (±0.00)}   & \textbf{0.86 (±0.00)}   & 0.53 (±0.05)                                  \\
        \bottomrule
        \label{tab:multi_venus}
        \end{tabularx}
        \end{table}
        \newline
        In general, results show how the performance of models utilizing multiple bands is superior to those using a single band. 
        Among the double-band combinations, B\(_5\)B\(_8\) and B\(_5\)B\(_{10}\) performed comparably, achieving precision values of 0.80 (±0.01) and 0.81 (±0.01), respectively. However, the B\(_5\)B\(_{12}\) combination exhibited a slightly higher recall of 0.87 (±0.00), leading to a superior $F_1$ score of 0.84 (±0.01). Moving to multi-band combinations, the triple-band setup B\(_3\)B\(_4\)B\(_7\) recorded the highest precision at 0.85 (±0.00), indicating a high degree of accuracy in detection while minimizing false positives. However, the 4-band combinations, such as B\(_3\)B\(_4\)B\(_7\)B\(_{11}\) and B\(_3\)B\(_5\)B\(_7\)B\(_{11}\), demonstrated a balanced performance across all metrics, with recall values of 0.87 (±0.00) and 0.88 (±0.00), respectively. Notably, B\(_3\)B\(_5\)B\(_7\)B\(_{11}\) achieved the highest metric values among all combinations, suggesting its effectiveness in comprehensive detection with fewer missed instances. 
        \newline
        However, it must be noted that although the application of multiple spectral bands yields performance improvements, these gains ($\approx$ 2\%) are not substantial enough to justify the onboard implementation of a co-registration technique.
        When a resource constrained environment is concerned, such as the one equipped onboard a cubesat, these improvements are traded with an increased power consumption and delays in processing time. 
       
\subsubsection{VDS2Raw}

        Complementary to the analysis performed for VENµS, we evaluate different spectral band combination for VDS2Raw, reporting the results in the Table \ref{tab:multi_sen}. The Table summarizes the detection metrics for various double and triple spectral band combinations. 
        \newline
        Among the double band combinations, \(B_3B_8\) shows slightly better performance in terms of Precision (0.64), Recall (0.85) and $F_1$ Score (0.74).
        \begin{table}[H]
        \centering
        \footnotesize
        \caption{Sentinel-2 (VDS2Raw) evaluation metrics computed for different spectral band combinations: Precision, Recall, $F_1$ score for detection, and MCC for classification.}
        \label{tab:multi_sen}
        \begin{tabularx}{\columnwidth}{p{3cm}p{3cm}p{3cm}p{3cm}|l}
        \toprule
        \textbf{Spectral Bands}             & \textbf{Precision}      & \textbf{Recall}         & \textbf{$F_1$}       & \textbf{MCC} \\ 
        \midrule
        \textbf{B$_8$}              & 0.59 (±0.08)            & \textbf{0.85 (±0.03)}            & 0.70 (±0.07)            & 0.33 (±0.01)               \\
        \textbf{B$_2$B$_4$}              & 0.55 (±0.04)            & 0.75 (±0.02)            & 0.63 (±0.03)            & 0.28 (±0.01)              \\
        \textbf{B$_2$B$_3$}              & 0.50 (±0.06)            & 0.77 (±0.03)            & 0.61 (±0.05)            & 0.33 (±0.04)              \\
        \textbf{B$_2$B$_8$}              & 0.56 (±0.05)         & 0.74 (±0.03)          & 0.64 (±0.04)            & 0.30 (±0.04)              \\
        \textbf{B$_3$B$_8$}              & 0.64 (±0.06)          & \textbf{0.85 (±0.02)}            & \textbf{0.74 (±0.05)}            & \textbf{0.39 (±0.07)}              \\
        \textbf{B$_4$B$_8$}              & 0.62 (±0.05)          & 0.83 (±0.02)            & 0.71 (±0.04)            & 0.38 (±0.02)              \\
        \textbf{B$_2$B$_4$B$_8$}         & 0.65 (±0.03)            & 0.76 (±0.02)            & 0.70 (±0.02)            & 0.32 (±0.02)         \\
        \textbf{B$_2$B$_3$B$_4$}         & 0.66 (±0.09)            & 0.80 (±0.08)            & 0.72 (±0.09)            & 0.34 (±0.06)                \\
        \textbf{B$_2$B$_3$B$_8$}         & \textbf{0.68 (±0.02)}   & 0.81 (±0.04)   & \textbf{0.74 (±0.02)}   & 0.34 (±0.03)              \\
        \bottomrule
        \end{tabularx}
        \end{table}
        The triple band combinations are in general better performing than the double combinations across all metrics. The combination \(B_2B_3B_8\) (blue, green, and near-infrared) consistently yields the best results, achieving the highest Precision (0.68), Recall (0.81), and $F_1$ Score (0.74), although the Recall rate is not the highest overall. This indicates that incorporating the near-infrared band \(B_8\) with the visible bands \(B_2\) and \(B_3\) enhances the model's ability to detect features more accurately. 
        The combination \(B_2B_4B_8\) (blue, red, and near-infrared) and \(B_2B_3B_4\) (blue, green, and red) also perform well, with $F_1$ Scores of 0.70 and 0.72, respectively, suggesting that adding the third spectral band improves overall performance. However, they are slightly less effective than \(B_2B_3B_8\), emphasizing the importance of the near-infrared band in improving detection accuracy.
        \newline
        In conclusion, adding a near-infrared band to visible bands resulted in improvements in performance, and our findings demonstrate that using a triple band combination, particularly the \(B_2B_3B_8\), is more effective for detection tasks than double band combinations. We argue that this is primarily due to the increased spectral diversity and complementary information provided. 
        Compared to the best single band, $B_8$, the improvement in performance evaluated in Precision is significant. Notwithstanding, the improvement is again negligible in terms of a more balanced metric such as $F_1$ score (0.74 vs. 0.70).
        Compared to VENµS, the higher performance improvement observed in a multispectral scenario suggests that leveraging broader spectral diversity can compensate for limited spatial resolution.

%% file: sections/05_towards.tex
\section{Onboard Proof-of-Concept}\label{DISCUSS}

The section first discusses the practical challenges of applying the model in a real-world scenario, particularly focusing on the onboard implementation, highlighting the practical results of the deployed system.
\newline
As demonstrated earlier, owing to the marginal improvements in multiple band combinations, thus we select a single band, i.e. $B_8$ for Sentinel-2 and the $B_{10}$ for VENµS. 
In deploying the models, we employed a cascaded approach for vessel detection and classification, i.e.,  after identifying regions of interest via the detector, the trained classifier was tasked with categorizing the vessels. This method was chosen to reduce the rate of misclassifications: a classifier trained end-to-end with the detector samples regions from the sea surface, thereby further complicating the task. While acknowledging the need for further investigation, we propose that future research should focus on refining and expanding this approach.
\newline
The model's robustness when facing different sensor conditions is evaluated in the first part of this section. 
Then, the practical implementation details are discussed, and finally, power and latency results are presented. 

\subsection{Impact of SNR and MTF}

\begin{figure}[htb]
        \centering
        \includegraphics[width=\linewidth]{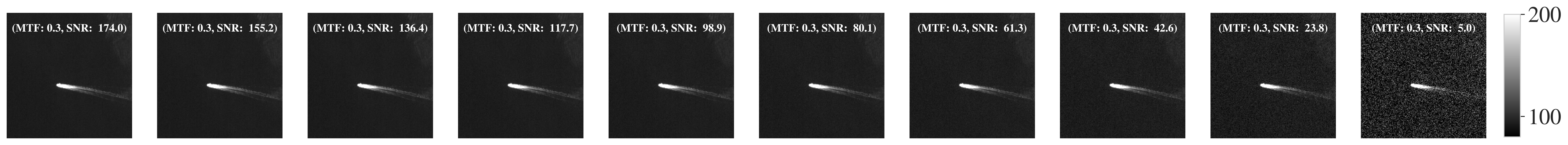}
        \includegraphics[width=\linewidth]{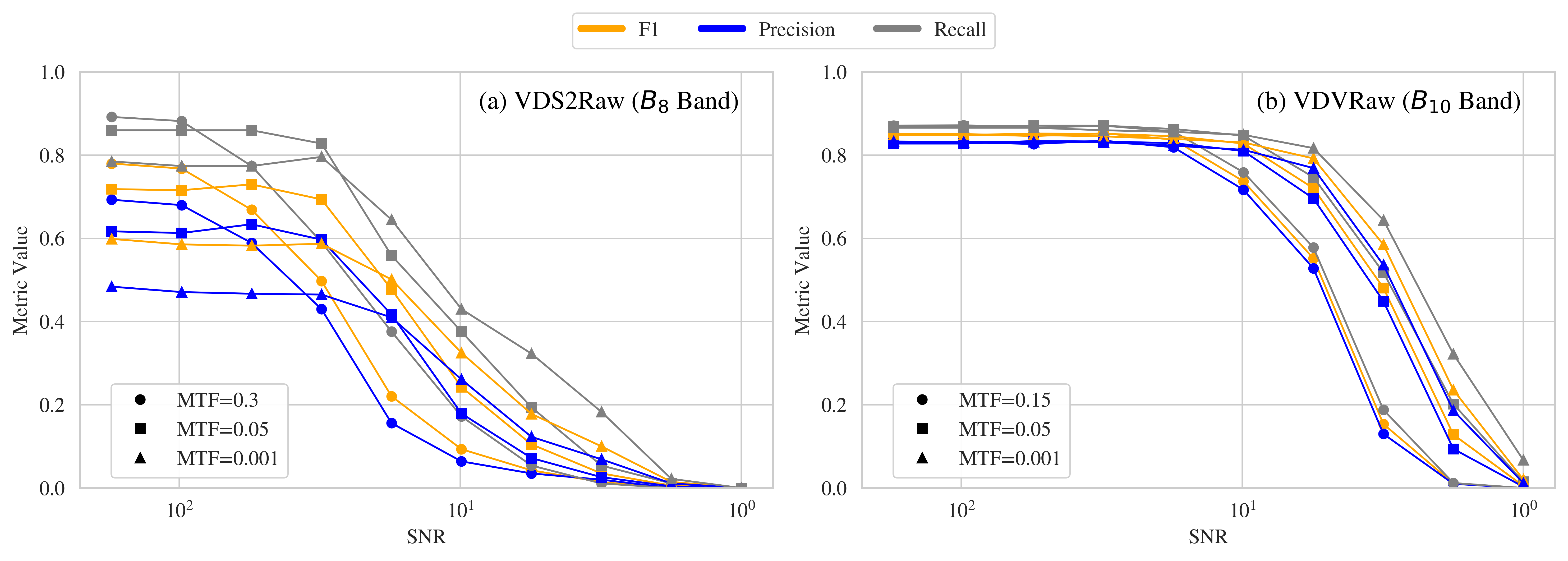}
        \includegraphics[width=\linewidth]{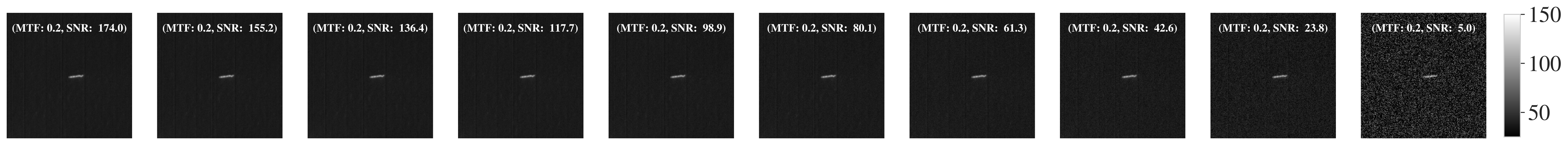}
        \caption{Detection performance metrics (precision, recall, and \( F_1 \) score) across varying levels of \gls{MTF} and \gls{SNR} for the datasets VDS2Raw (left panel) and VDVRaw (right panel). The top and bottom panels show the impact of different \gls{SNR} values on Sentinel-2 and VENµS raw tiles, respectively.}
        \label{fig:MTFSNR}
    \end{figure}

    Critical insights can be drawn when benchmarking models under different sensor characteristics, specifically the \gls{MTF} and \gls{SNR}. 
    This information is crucial as the onboard data distribution can shift significantly from the one faced during training, as remarked in \cite{longepe2024simulation}. 
    We evaluate the effect of this domain gap by analysing the model behaviour against different sensor working conditions. 
    \newline
    The optical system can be modelled using the \gls{MTF} and \gls{SNR} to simulate the sensor's spatial response and noise levels. The \gls{MTF} aims to replicate the spatial domain response of the image (i.e., emulating the signal recorded by the instrument detector element), which is closely associated with the instrument characteristics, such as pixel size and system F-number. The \gls{MTF} filter is thus represented as a two-dimensional kernel with independent across- and along-track parameterizations. By utilizing key instrument parameters (mainly the pixel pitch that determines the sampling frequency and the F-number that determines the cutoff frequency for each band), a two-dimensional \gls{MTF} kernel of size $n \times n$ is computed \cite{longepe2024simulation}. The corresponding \gls{PSF} kernel is obtained as the inverse Fourier transform of the \gls{MTF} kernel, and this PSF kernel is convolved with the input image using a sliding-window approach.
    \newline
    Let $M(B_i)$ the \gls{MTF} amplitude at the Nyquist frequency for each spectral band $B_i$, the \gls{MTF} function (expressed in terms of normalized frequency with respect to the Nyquist) can be represented as:
    \begin{equation}
        MTF(B_i, k) = e^{\ln{M(B_i)} \cdot k^2}
    \end{equation}
    The $M(B_i)$ are known for Sentinel-2 ($M_{S2}(B_i) \approx 0.3$ \cite{gascon}), and for VENµS ($M_{V}(B_{10}) \approx 0.15$ \cite{gamet2019measuring}), therefore we can perform a deconvolution of the input image with the source PSF followed by a convolution with the target PSF as in the simulator tool of \cite{longepe2024simulation}, with a single step adopting the Gaussian hypothesis.
    \newline
    Concerning the noise simulation, an additional noise term ($N$) is imposed to the raw images to account for \gls{SNR} conditions, following again the formulation by \cite{longepe2024simulation}:
    \begin{equation} \label{eq:SNR}
        N(i, j, b_n) = \frac{L_{ref}(b_n)}{SNR(b_n)} \cdot \mathcal{N}(0, 1),
    \end{equation}
    where $\mathcal{N}(0, 1)$ is a Gaussian variable with zero mean and unit standard deviation, and $L_{ref}$ the reference radiance. 
    Notably, the Equation (\ref{eq:SNR}) must be modified from the work of \cite{longepe2024simulation} to address raw data since the information is stored as \gls{DN}s. Therefore, instead of a reference radiance, we adopt a $DN_{ref}$ of 100 for both VENµS and Sentinel-2 imagery.
    We impose this additional noise term on the existing noise, starting from a \gls{SNR} value of 174 (theoretical \gls{SNR} ratio for Sentinel-2 at the $B_8$ band); we assume the same initial value also for VENµS. 
    \newline
    Under the hypothesis of this modelling framework, we apply varying levels of \gls{SNR} and \gls{MTF} to evaluate the model's robustness and generality. 
    The results, displayed in Figure \ref{fig:MTFSNR}, illustrate the detection metrics: precision, recall, and $F_1$ score. As shown in the graphs, the left panel displays results for the VDS2Raw dataset, while the right panel shows results for VDVRaw
    The results indicate that lower \gls{MTF} values---leading to increased image blurring---decrease the detection performance, with precision metric being the most affected among the three.
    \newline
    At higher \gls{SNR} rates, evidence also shows that the \gls{S-2} model is much more influenced by \gls{MTF} variations, mainly due to the lower spatial resolution. 
    An interesting behaviour is noted for both models: lower \gls{MTF} values shift the plateau from which model performance begins to decline rapidly. This can be attributed to the fact that lower \gls{MTF} values act as a median filter, thus mitigating the noise and thereby reducing the number of false alarms. In summary, for relatively low \gls{SNR} values, having a lower \gls{MTF} is advantageous despite the overall reduced performance.
    \newline
    Regarding the \gls{SNR}, lower values, which correspond to increased noise (as illustrated in the top and bottom panels of Figure \ref{fig:MTFSNR} for Sentinel-2 and VENµS imagery, respectively), lead to reduced detector performance.
    The results highlight again that VDVRaw is more resilient to noise due to the higher spatial resolution of the sensor, whereas the model trained on VDS2Raw demonstrates degraded performance. 
    
    \subsection{Onboard Implementation}
    
\begin{figure}[tb]
        \centering
        \includegraphics[width=0.5\linewidth]{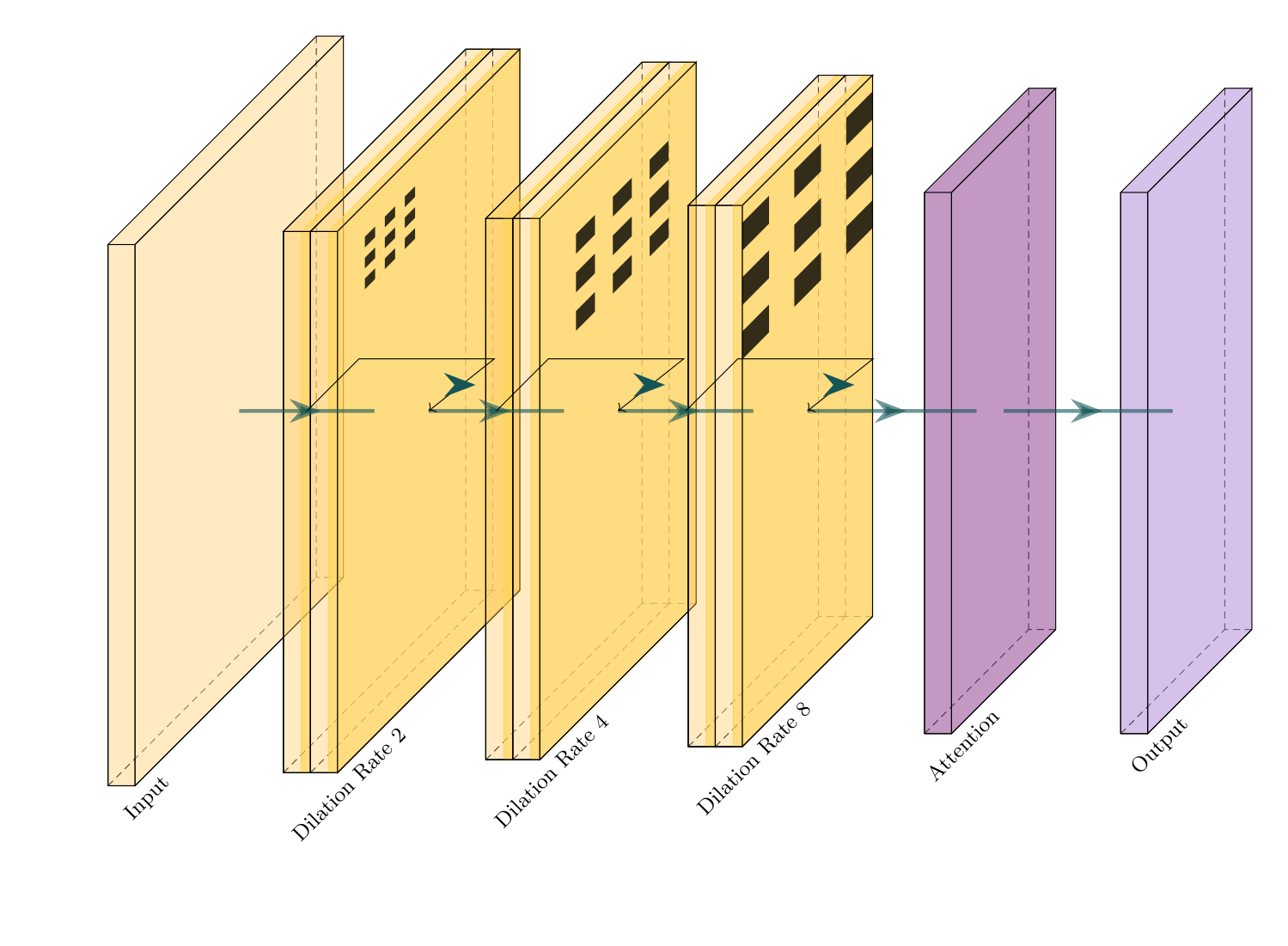}
        \caption{Proposed \gls{RADC} module integrates skip connections with dilated convolutions at multiple dilation scales to enhance feature extraction across various receptive fields. The module also incorporates a channel attention mechanism to dynamically re-weight the channels on the last layer of the module.}
        \label{fig:RADC}
    \end{figure}

    Benchmarking tests were conducted on a Raspberry Pi 4B (Quad core Cortex-A72 ARM v8 64-bit SoC @ 1.5GHz), a popular low-power device, coupled with the Intel\textsuperscript{\textregistered} \gls{NCS2} connected through USB bus. The \gls{NCS2} features the Intel\textsuperscript{\textregistered} Movidius\texttrademark{} Myriad\texttrademark{} X \gls{VPU}. 
    This chip is designed to accelerate deep learning tasks at the edge, featuring 16 programmable 128-bit VLIW Vector Processors and can execute over 4 trillion operations per second (TOPS). 
    \newline
    To enable onboard deployment, several modifications were made to the detection model. The input dimensions were reduced to optimise memory usage, aligning with the memory constraints of the \gls{VPU}. 
    As this \gls{VPU} does not natively support deformable convolutions, the star-shaped deformable convolutions were replaced with a custom-developed module.
    More in-depth, we introduce the \gls{RADC} module. 
    The \gls{RADC} module, drawn in Figure \ref{fig:RADC}, integrates skip connections with dilated convolutions at multiple dilation scales.
    \newline
    For each dilated convolutional block with dilation $d_i$, the output $\mathbf{X}_{i}$ is given by:
    \begin{equation}
    \mathbf{X}_{i} = \text{ReLU}(\mathbf{X} *_{d_i} \mathbf{W}_{i} + \mathbf{b}_{i}) + \mathbf{X}
    \end{equation}
    where $\mathbf{X}$ is the input tensor, $*_{d_i}$ denotes the dilated convolution with dilation rate $d_i$, $\mathbf{W}_{i}$ represents the weights of the $i$-th dilated convolutional layer, $\mathbf{b}_{i}$ is the bias of the $i$-th dilated convolutional layer, and $\text{ReLU}(\cdot)$ is the Rectified Linear Unit activation function.
    Subsequently, the \gls{CAM} computes an attention map $\mathbf{A}$ that re-weights the channels of the output:
    \begin{equation}
    \mathbf{A} = \sigma(\mathbf{X}_3 * \mathbf{W}_{\text{att}})
    \end{equation}
    where $\sigma(\cdot)$ is the sigmoid activation function, and $\mathbf{W}_{\text{att}}$ represents the weights of the attention layer.
    The attention map $\mathbf{A}$ is applied to the output of the last dilated convolutional block:
    \begin{equation}
    \mathbf{X}_{\text{att}} = \mathbf{X}_3 \odot \mathbf{A}
    \end{equation}
    The different dilation rates ($d_i$ where $i \in $   [2,4,8]) favour learning multi-scale feature representation by greatly enlarging the receptive fields at different levels, thereby helping the modelling of complex sea surface distributions.
    The CAM layer further refines the feature maps, enhancing the model's ability to focus on the most important channels for vessels and the sea surface.
    \newline
    This new version of the model built on top of VFNet--termed RADCNet--is then trained using automatic mixed-precision technique. 
    Upon completing the training phase, the model is converted into its Intermediate Representation (IR) using the OpenVINO framework, utilizing the $16$-bit floating-point format. This conversion is necessitated by the \gls{VPU}'s limitation to $16$-bit floating-point precision. An identical procedure is applied to the classification model, ensuring both models are compiled in IR with the same numerical precision, thereby adhering to the hardware constraints imposed by the \gls{VPU} and optimizing performance for deployment.
    \newline
    RADCNet closely matches the original VFNet across all metrics, maintaining high precision, recall, and $F_1$ scores, as shown in Table \ref{tab:performance_metrics_comparison}.
    \newline
    In terms of parameter count, RADCNet exhibits a modest increase of approximately $+3$ million parameters, primarily attributable to the introduction of the \gls{RADC} module. However, in terms of throughput, RADCNet substantially surpasses VFNet, achieving a frame rate of $18$ images per second compared to VFNet's $14$ images per second. Latency measurements were conducted over $2000$ iterations, with $5$ warmup iterations, on the Linux system detailed in Table~\ref{tab:model_evaluation}. These results highlight the significant influence of memory access costs and demonstrate RADCNet's competitive performance, underscoring its suitability for deployment in resource-constrained environments.
    \newline
    \begin{table}[!h]
    \centering
    \caption{Comparison of Precision, Recall, $F_1$ score, number of parameters, and FPS between VFNet and RADCNet models on VDS2Raw ($B_8$ band) and VDVRaw ($B_{10}$ band) datasets. Despite being quantized, RADCNet shows minimal performance degradation. [Benchmark computed using 2000 iterations with a warmup of 5]}
    \label{tab:performance_metrics_comparison}
    \begin{tabularx}{\textwidth}{@{}l*{6}{>{\centering\arraybackslash}X}@{}}
    \toprule
    \textbf{Model}  & \textbf{Dataset} & \textbf{Precision}       & \textbf{Recall}          & \textbf{$F_1$}          & \textbf{Params (M)} & \textbf{FPS (img/s)} \\ \midrule
    VFNet           & VDS2Raw          & 0.59 (± 0.08)            & \textbf{0.85 (± 0.03)}   & 0.70 (± 0.07)           & \textbf{19.68}      & 14.9                 \\
    RADCNet         & VDS2Raw          & \textbf{0.63 (± 0.05)}   & 0.81 (± 0.04)            & \textbf{0.71 (± 0.04)}  & 22.04               & \textbf{18.0}        \\ \midrule
    VFNet           & VDVRaw           & \textbf{0.82 (± 0.01)}   & \textbf{0.87 (± 0.00)}   & \textbf{0.85 (± 0.00)}  & \textbf{19.68}      & 14.9                 \\
    RADCNet         & VDVRaw           & 0.80 (± 0.02)            & 0.84 (± 0.03)            & 0.82 (± 0.02)           & 22.04               & \textbf{18.0}        \\ \bottomrule
    \end{tabularx}
    \end{table}

    \subsection{Power and Latency Benchmarks}
    
\begin{figure}[b]
        \centering
        \includegraphics[width=\linewidth]{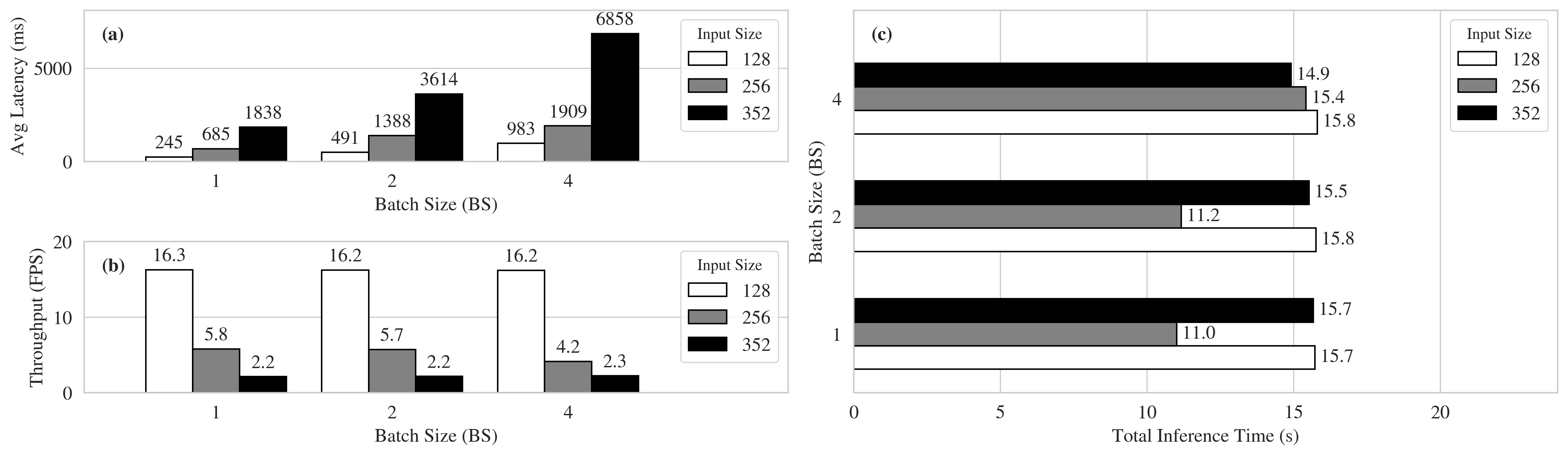}
        \caption{Benchmark of detection model compiled using OpenVINO: graph (a) presents the average latency (ms), graph (b) illustrates the average throughput (FPS), and graph (c) displays the total inference time for an image with dimensions 2048 $\times$ 2048 pixels.}
        \label{fig:benchmark}
    \end{figure}

\begin{figure}[htb]
        \centering
        \includegraphics[width=\linewidth]{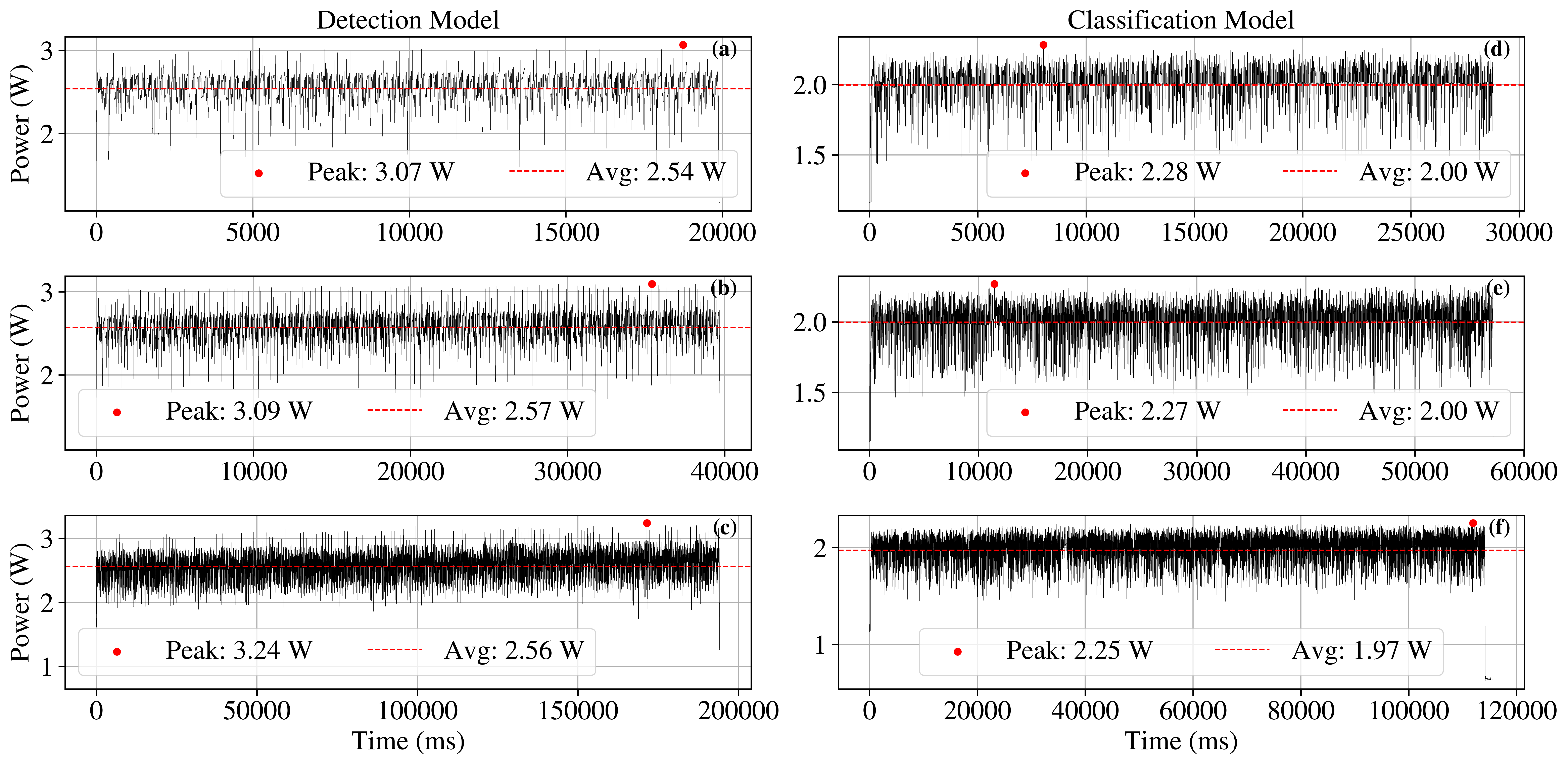}
        \caption{Graph displaying the power consumption of the Intel\textsuperscript{\textregistered} Neural Compute Stick 2 (NCS2) connected through USB bus to the Raspberry Pi 4B. The device performs separate inferences using the detection model (graphs (a), (b), and (c), having an input shape of 256x256 with batch size 1 and 2, and 352x352 with batch size 4, respectively) and the classification model (graphs (d), (e) and (f), having an input shape of 64x64 with batch size 8, 16, and 32, respectively) that require lower total inference time (Figure \ref{fig:benchmark}). In the proposed setup, the NCS2 consumes 0.66W in its idle state, data sampling is performed with a $\Delta t = 10\ \text{ms}$ using the FNIRSI FNB58 USB tester, and the number of inferences selected for the detection and classification models is 100 and 1000, respectively.}
        \label{fig:power}
    \end{figure}

    This subsection presents the timing and power consumption results of the on-device tests. Latency was first evaluated on single models, then the entire processing chain, including pre-processing (tiling, normalization), and inference, encompassing both detection and classification has been benchmarked. 
    \newline
    Figure \ref{fig:benchmark} presents the evaluation conducted on the detection model, whereas Figure \ref{fig:power} displays the power consumption of the edge-AI hardware chosen performing separate detection and classification tasks.
    \newline
    Several batch size dimensions have been considered along with different input sizes. The results are averaged over one thousand iterations using the benchmark tool of OpenVINO. The average inference time is shown in Figure \ref{fig:benchmark}(a). Additionally, the throughput has been estimated and reported in Figure \ref{fig:benchmark}(b). In the end, the graph of Figure \ref{fig:benchmark}(c) considers the input size of 2048 $\times$ 2048 pixel image and applies the \gls{SAHI} approach \cite{sahi} to perform a complete inference using smaller tiles.
    \newline
    Regarding power consumption, Figure \ref{fig:power} illustrates the power consumption profile of the Intel\textsuperscript{\textregistered} \gls{NCS}2 connected via USB to the Raspberry Pi 4B. The device executes inference tasks using both detection and classification models. Specifically, subplots (a), (b), and (c) display the power consumption for the detection model with input dimensions of 256 $\times$ 256 for batch sizes of 1 and 2, and 352 $\times$ 352 for a batch size of 4, respectively. 
    Instead, subplots (d), (e), and (f) show the corresponding results for the classification model with input dimensions of 64 $\times$ 64 for batch sizes of 8, 16, and 32, respectively. These models exhibit lower total inference times as depicted in Figure \ref{fig:benchmark}. 
    In the proposed configuration, the NCS2 operates with a baseline idle power consumption of 0.66W. Power measurements are sampled at intervals of $\Delta t = 10\ \text{ms}$ using the FNIRSI FNB58 USB tester. The number of inference operations conducted is set to 100 for the detection models and 1000 for the classification models.
    \newline
    Regarding the classification task, we repeat the same steps for the classifier model, obtaining the latency and throughput values reported in Table \ref{tab:tabella_bench}. The results indicate that the throughput remained relatively stable across batch sizes, ranging from 306.61 \gls{FPS} (batch size 2) to 329.99 \gls{FPS} (batch size 16). As expected, the average inference time increased with larger batch sizes, from 12.70 ms at batch size 1 to 388.22 ms at batch size 32. These findings suggest that the optimal balance between throughput and inference time occurs at batch size 4, where a high throughput (319 \gls{FPS}) is achieved with manageable latency (50ms).
    \newline
\begin{table}[ht]
\centering
\caption{Throughput, average, minimum and maximum inference time of the classification model deployed with the OpenVINO framework, prompted for different batch sizes.}\label{tab:tabella_bench}
\begin{tabular}{@{}ccccc@{}}
\toprule
\textbf{Batch Size} & \textbf{Throughput (FPS)} & \textbf{Avg Inference Time (ms)} & \textbf{Min Inference Time (ms)} & \textbf{Max Inference Time (ms)} \\ \midrule
1          & 308.53           & 12.70                   & 9.47             & 17.90            \\
2          & 306.61           & 25.76                   & 14.99            & 30.82            \\
4          & 319.22           & 49.70                   & 26.66            & 53.86            \\
8          & 328.04           & 97.01                   & 48.26            & 103.63           \\
16         & 329.99           & 193.30                  & 95.27            & 200.12           \\
32         & 328.70           & 388.22                  & 202.70           & 408.18           \\ \bottomrule
\end{tabular}
\end{table}
    In the end of the manuscript, we present an end-to-end demonstration of our system, processing a complete acquisition under a worst-case scenario involving 200 vessels per acquisition, with results averaged over 10 iterations. The pre-processing phase is highly efficient, with an average duration of 0.49 seconds, suggesting that tasks such as normalization and resizing are not the primary computational bottlenecks. Conversely, the detection inference stage, tasked with identifying vessel locations, is the most time-consuming step, with an average execution time of 22.97 seconds, underscoring the complexity of the detection process. The classification phase, where detected vessels are categorized, is significantly faster, averaging 1.77 seconds, reflecting the optimized nature of this step.
    \newline
    These performance metrics indicate that the detection stage consumes the majority of computational resources. Notably, our system's execution time is approximately double that of comparable benchmarks, which we attribute to our Python-based implementation as opposed to the more optimized C++ implementations used in the benchmark tools. 

%% file: sections/06_conclusion.tex
\section{Conclusions and Future work}\label{CONCLUSIONS}

The paper showcased the capability of vessel identification onboard satellites in an end-to-end fashion, starting from raw, unprocessed, multispectral imagery. For this purpose, two novel datasets have been compiled, encompassing two distinct geographic regions captured by different sensors with unique spatial and radiometric characteristics. The diversity, also reflected in terms of noise and artefacts, allowed us to explore the performance of our detection models across varying conditions.
\newline
Through a comprehensive statistical analysis, including feature and metric estimations, we identified the most effective spectral bands for vessel detection, i.e., $B_8$ and $B_{10}$ for Sentinel-2 and VENµS, respectively. The spectral responses of these two bands are closely aligned, covering similar portions of the spectrum. These findings are significant, as they not only enhance our understanding of the optimal bands for this specific task but also demonstrate the robustness of our approach in adapting to different sensor configurations and environmental contexts. 
Additionally, our study indicates that the advantages of using multiple spectral bands, particularly in improving detection accuracy, are negligible in the context of onboard-constrained resources.
\newline
Finally, with a bare minimal pre-processing, we achieved onboard execution in under 26 seconds on a representative onboard edge-AI device (Raspberry Pi 4B + NCS2) with a power consumption below 2W registered on the NCS2, showcasing the feasibility of near real-time vessel detection and classification.
Notwithstanding, it must be pointed out that, at the time of writing this manuscript, other models have been released, such as YOLOv10 \cite{wang2024yolov10}, and Real-Time DEtection TRansformer (RT-DETR) \cite{zhao2024detrs}, which exhibit exceptional performance in detection tasks.
These models benefit from end-to-end capabilities for detecting objects thanks to one-to-one prediction assignments, thus eliminating the need for \gls{NMS} component. Given the promising results demonstrated by these models, their application and performance evaluation in the context of our research would be valuable. However, due to the rapid advancements in this field and the scope of our current study, the analysis of these more representative models is deferred to future work.
\newline
Although results demonstrated the detection phase is the computational bottleneck, the adopted cascaded approach can be further refined by directly integrating the classifier in the detection heads. This modification could streamline the workflow, allowing for a reduction in latency and overall better resource management. In addition, findings suggest that future work aiming for enhanced efficiency should consider leveraging C++ for improved resource utilization.
\newline
The AIS records have not been fully exploited. Future work should focus on identifying the classes that are most challenging to detect. This will enable to optimize our models for improved efficiency in targeting those specific classes.
\newline
Finally, noise appears to affect model performance significantly. Hence, future research should explore integrating advanced denoising techniques to mitigate the impact of noise. Incorporating such techniques could enhance the robustness of the models, especially in real-time operational settings, and lead to more accurate and reliable vessel detection and classification. We believe that developing models resilient to variations in \gls{MTF} and \gls{SNR} represents a promising direction for future research addressing domain gap challenges.

\section*{Data Availability}
\begin{itemize}
    \item The VDS2Raw (v2) dataset, which is integral to our research, is publicly accessible for further exploration and utilization. Researchers and practitioners can download the dataset from the following link: \textit{10.5281/zenodo.13889073}. 

    \item The VDVRaw dataset is publicly accessible for further exploration and utilization. The dataset is accessible at the following link: \textit{10.5281/zenodo.13897485}. 

    \item Both classification datasets are publicly accessible for further exploration and utilization at the following link: \textit{10.5281/zenodo.14007820}.
\end{itemize}

\subsection*{Declaration of Generative AI}
During the preparation of this work the authors used GPT-4o in order to improve the readability of some sentences. After using this tool/service, the authors reviewed and edited the content as needed and took full responsibility for the content of the publication.